\documentclass[acmtog]{acmart}
\usepackage{xspace}
\usepackage{enumitem}
\usepackage{algorithm}
\usepackage{algpseudocode}
\usepackage{array}    
\usepackage{booktabs} 

\usepackage{booktabs}
\usepackage{array}
\usepackage[percent]{overpic}
\usepackage[table]{xcolor}
\usepackage{colortbl}
\definecolor{bestcolor}{rgb}{0.7,0.9,0.99}     
\definecolor{secondcolor}{rgb}{0.88,0.96,1.0}   
\newcolumntype{C}[1]{>{\centering\arraybackslash}m{#1}} 

\usepackage{xcolor}
\usepackage{pifont}   

\definecolor{markgreen}{RGB}{0,160,80}
\definecolor{markred}{RGB}{220,50,32}
\definecolor{markorange}{RGB}{230,150,0} 

\newcommand{\cmark}{\textcolor{markgreen}{\ding{51}}} 
\newcommand{\xmark}{\textcolor{markred}{\ding{55}}}   
\newcommand{\pmark}{\textcolor{markorange}{\ding{108}}} 

\AtBeginDocument{%
  }

\acmJournal{TOG}                                                              
  \setcopyright{acmlicensed}
  \copyrightyear{2026}                                                          
  \acmYear{2026}                                            
  \acmDOI{XXXXXXX.XXXXXXX}   
  \acmVolume{45}                                                                
  \acmNumber{4}                                                                 
  \acmArticle{XXX}           
  \acmMonth{7}                                                                  
                                                                                
  \acmConference[SIGGRAPH '26]{ACM SIGGRAPH 2026 Conference Papers}{July 19--23,
   2026}{Los Angeles, CA, USA}                                                  
  \acmBooktitle{ACM SIGGRAPH 2026 Conference Papers (SIGGRAPH '26), July 19--23,
   2026, Los Angeles, CA, USA}                                                  
  \acmISBN{978-1-XXXX-XXXX-X/26/07}   

\begin{document}


\newcommand{\Frank}[1]{{\color{black}{#1}}}

\newcommand{\Lucky}[1]{{\color{black}{#1}}}

\newcommand{\name}{SAD\xspace}

\title{Soft Anisotropic Diagrams for Differentiable Image Representation}


\author{Laki Iinbor}
\email{lacoz@icloud.com}
\orcid{0009-0001-7499-6416}
\affiliation{%
  \institution{Independent Researcher}
  \country{USA}
}
\thanks{Visiting research assistant at MIT CDFG}

\author{Zhiyang Dou$^\dagger$}
\email{frankdou@mit.edu}
\orcid{0000-0003-0186-8269}
\affiliation{%
  \institution{MIT}
  \state{Massachusetts}
  \country{USA}
}

\author{Wojciech Matusik$^\dagger$}
\email{wojciech@csail.mit.edu}
\orcid{0000-0003-0212-5643}
\affiliation{%
  \institution{MIT}
  \state{Massachusetts}
  \country{USA}
}
  \thanks{$\dagger$ Joint Last Author}

\renewcommand{\shortauthors}{Iinbor et al.}

\begin{abstract}
We introduce Soft Anisotropic Diagrams~(\name), an explicit and differentiable image representation parameterized by a set of adaptive sites in the image plane. In \name, each site specifies an anisotropic metric and an additively weighted distance score, and we compute pixel colors as a softmax blend over a small per-pixel top-$K$ subset of sites. We induce a soft anisotropic additively weighted Voronoi partition (i.e., an Apollonius diagram) with learnable per-site temperatures, preserving informative gradients while allowing clear, content-aligned boundaries and explicit ownership. Such a formulation enables efficient rendering by maintaining a per-query top-$K$ map that approximates nearest neighbors under the same shading score, allowing GPU-friendly, fixed-size local computation. We update this list using our top-$K$ propagation scheme inspired by jump flooding, augmented with stochastic injection to provide probabilistic global coverage. Training follows a GPU-first pipeline with gradient-weighted initialization, Adam optimization, and adaptive budget control through densification and pruning. Across standard benchmarks, \name consistently outperforms Image-GS and Instant-NGP at matched bitrate. On Kodak, \name reaches 46.0 dB PSNR with 2.2 s encoding time (vs. 28 s for Image-GS), and delivers 4–19$\times$ end-to-end training speedups over state-of-the-art baselines. We demonstrate the effectiveness of \name by showcasing the seamless integration with differentiable pipelines for forward and inverse problems, efficiency of fast random access, and compact storage. You can find the code here: \url{https://luckyiyi.github.io/SAD}.
\end{abstract}


  \begin{teaserfigure}
    \includegraphics[width=\textwidth]{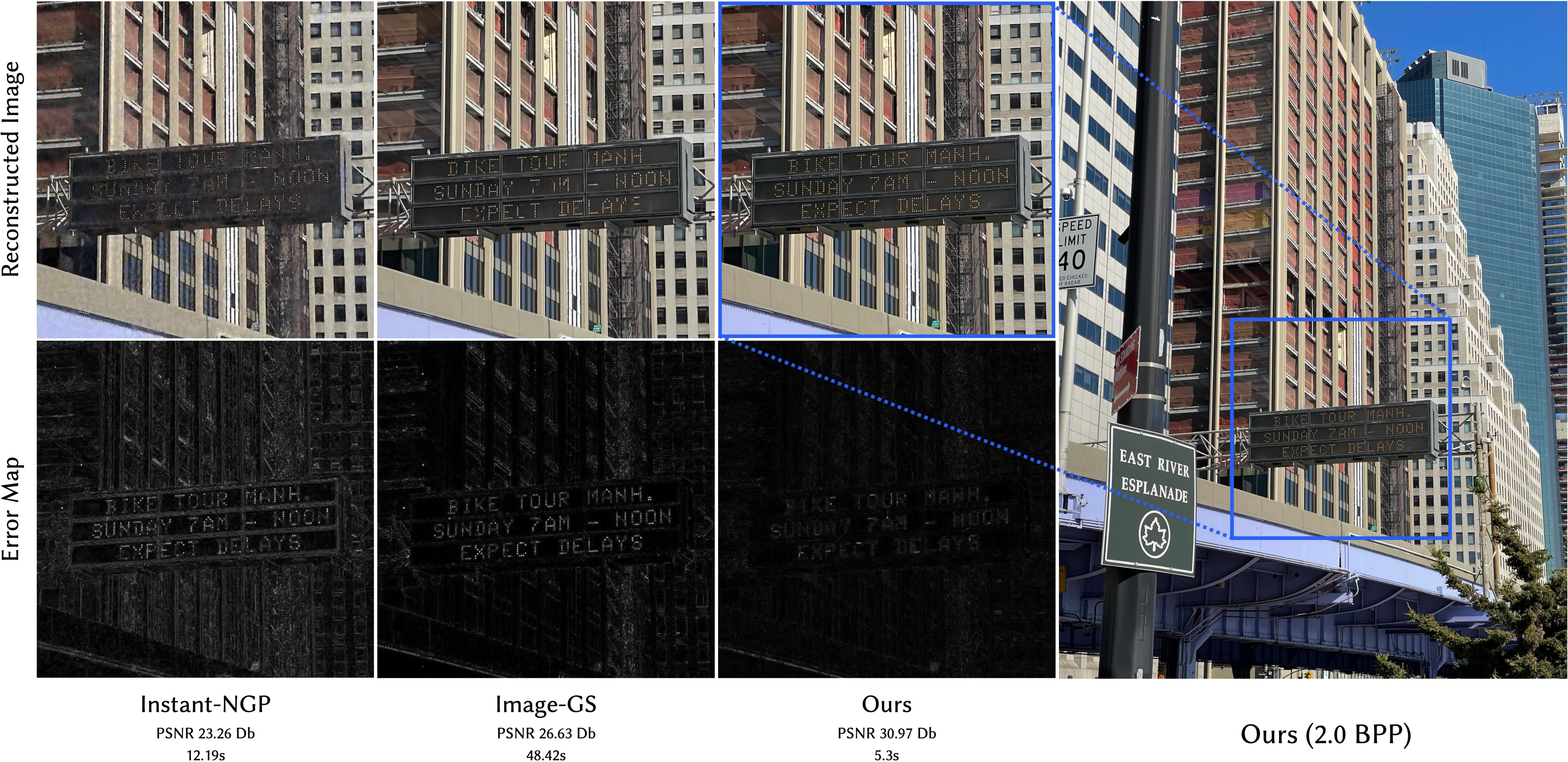}
    \vspace{-6mm}
    \caption{\name achieves superior
    reconstruction quality and remarkably higher training speed across different compression rates and resolutions. At 2.0 BPP (shown here), our method delivers 30.97 dB PSNR in 5.3s, compared to Image-GS~\cite{imagegs2025} (26.63 dB, 48.42s) and Instant-NGP~\cite{mueller2022instant} (23.26 dB, 12.19s). Error maps (bottom row, 2× scaled) reveal significantly lower error on sharp edges and structural boundaries. Here, zoom insets demonstrate accurate reconstruction of fine details through our temperature-controlled soft Apollonius partition.}
    \Description{Comparison of three image reconstruction methods demonstrating \name' superior speed and quality.}
    \label{fig:teaser}
  \end{teaserfigure}

\received{22 Jan 2026}
\received[revised]{19 Apr 2026}
\received[accepted]{20 Apr 2026}


\maketitle

\section{Introduction}

Efficient, compact, and differentiable image representations~\cite{balle2016end, imagegs2025, gaussianimage2024, mueller2022instant,sitzmann2020implicit,ulyanov2018deep, wang2025fast,instantgi2025,zhu2025large} that are fast to fit and evaluate are increasingly important for a wide range of applications in vision and graphics, including random-access compression, continuous decoders in generative models, adaptive-resolution evaluation and differentiable resampling for analysis as well as editing, and as compact priors for different inverse problems.

Many efforts~\cite{tancik2020fourier, saragadam2023wire, sitzmann2020implicit,dupont2021coin} adopt implicit neural representations for their flexibility and expressiveness. However, they typically lack explicit structure: spatial ownership is not directly represented, pruning and localized reallocation are non-trivial, and controlling representation budget often relies on indirect heuristics. Explicit splat-based representations~\cite{imagegs2025,gaussianimage2024} improve content adaptivity and can be efficient to query, yet kernel overlap blurs per-pixel responsibility and complicates pruning and budget control—especially when sharp discontinuities must be represented without excessive overlap. Beyond structural limitations, a persistent obstacle is \emph{encoding} cost: fitting a compact representation while maintaining high quality is often orders of magnitude slower than evaluating it. This gap hinders high-throughput and resource-constrained settings—such as large dataset encoding, video representation, interactive workflows, and general-purpose compression on consumer devices. 

{These considerations motivate a design choice: we favor a representation in which each pixel depends on a small, fixed number of primitives with an interpretable neighborhood structure, i.e., explicit spatial locality and adjacency.} Such locality provides (i) predictable query cost, (ii) localized responsibilities that facilitate pruning and densification, and (iii) the ability to align discontinuities with image content rather than smearing them through kernel overlap. Crucially, locality maps naturally to GPU execution: constant-size per-pixel neighborhoods enable regular, bandwidth-friendly kernels (coalesced reads and shared-memory reductions) and propagation-style updates, avoiding overlap-driven culling pipelines and contention-heavy gradient accumulation.

Motivated by these considerations, we introduce \name, an explicit, differentiable image representation that replaces kernel overlap with a \emph{learnable soft partition} of the 2D image plane. We parameterize the image with adaptive anisotropic sites and render each pixel via a softmax blend over its top-$K$ sites, using a site-dependent distance score under a learnable anisotropic metric.
This fixed-$K$ per-pixel neighborhood makes both rendering and fitting GPU-friendly, enabling regular local kernels and efficient propagation-style updates.
This yields a partition of unity with dense gradients while making ownership explicit. Learnable per-site temperatures can then sharpen the partition into crisp, content-aligned boundaries and expose adjacency structure useful for analysis, compression, and downstream learning. Geometrically, \name can be viewed as a \emph{generalized Voronoi} construction~\cite{voronoi1908nouvelles,aurenhammer1991voronoi} (also known as an Apollonius diagram\footnote{In this paper, we use the terms \emph{Apollonius diagram} and \emph{additively weighted Voronoi diagram} interchangeably.}): different choices of the per-site score recover familiar tessellations. In our instantiation, we adopt an \emph{additively weighted} form (a per-site radius offset)~\cite{emiris2006predicates}, which integrates cleanly with our anisotropic norm, provides an intuitive ``influence radius'' per site, and---through the softmax relaxation---supports a soft-to-sharp transition controlled by temperature. {In this paper, although we use the term ``diagram'' to emphasize the underlying Apollonius-style score and induced neighborhood structure, the rendered representation is a soft partition of unity rather than a hard nearest-site tessellation.} 

\name avoids evaluating all $N$ sites per pixel by maintaining a fixed-size per-pixel top-$K$ list ($K{=}8$) under the \emph{same} score used for shading---an approximate $K$-order Voronoi query. We update this list with our top-$K$ propagation scheme inspired by Jump Flooding Algorithm (JFA)~\cite{rong2006jumpflooding}: temporal warm starts, spatial propagation from a small fixed neighborhood (self + 4 neighbors), and stochastic injection for probabilistic global coverage. This yields fixed per-pixel compute, i.e., $O(P\!\cdot\!K)$ work per pass plus $O(P\!\cdot\!K)$ for rendering or gradient computation with small constants (with $P{=}H\!\cdot\!W$ pixels), rather than overlap-dependent scanning and per-iteration rebuilds of global acceleration structures; in practice, the resulting kernels are highly regular and bandwidth-efficient while reducing the atomic contention during backpropagation that hinders the performance of methods such as Image-GS~\cite{imagegs2025}.

Our experiments show that \name improves the quality--efficiency trade-off of compact image representations, reducing per-instance encoding cost while improving rate--distortion quality. On the Image-GS benchmark, \name consistently outperforms Image-GS and Instant-NGP at matched bitrate (e.g., 37.86~dB at 0.5~{bits-per-pixel~(BPP)} in Table~\ref{tab:image_gs_results}).
On Kodak~\cite{kodak} with $N{=}50{,}000$ primitives, \name achieves 46.00~dB while cutting encoding time from 28~s to 2.2~s under the same protocol (Table~\ref{tab:kodak_compression}). 
Moreover, \name is substantially faster to fit in wall-clock time (up to 19$\times$ over Image-GS), and reaches visually clean reconstructions much earlier in optimization (e.g., 5~s vs.\ 48~s in Figure~\ref{fig:teaser}). Beyond reconstruction, explicit ownership and induced adjacency make \name a convenient primitive for downstream tasks that benefit from local control and hard spatial constraints. We demonstrate its application in differentiable PDE solving on irregular domains with hard boundary constraints. We further perform several ablation studies to validate our design choices. In summary, we make the following contributions:

\begin{itemize}[leftmargin=1.2em]
  \item We propose \name, a soft anisotropic additively weighted (Apollonius-style) partition-of-unity image model with learnable per-site temperatures, enabling sharp content-aligned boundaries and explicit ownership.
  \item We develop a GPU-friendly top-$K$ propagation algorithm that maintains per-pixel top-$K$ lists via reuse, jump-flood propagation, and global probing, yielding constant per-pixel query cost.
  \item We present a GPU-first optimization and budget-control pipeline with adaptive densification and removal-delta pruning, together with efficient gradient accumulation.
\end{itemize}

\section{Related Work}
\subsection{Image Representations and Differentiable Partitions} \label{sec:rw_partitions} 
The field of neural image representation and compression has a rich history, spanning learned transform coding and end-to-end optimization~\cite{balle2016end, cheng2020learned, theis2017lossy} as well as implicit, coordinate-based signal models~\cite{dupont2021coin, tancik2020fourier, sitzmann2020implicit, ladune2023cool, karnewar2022relu, martel2021acorn, vaidyanathan2023random, mueller2022instant}. Recently, point- or splat-based primitives have also emerged as compact image representations~\cite{imagegs2025, gaussianimage2024}. In the following, we focus on recent implicit neural signal representations and on point-based and splat-based rendering techniques.

\paragraph{Implicit neural signal representations.}

Implicit coordinate-based networks represent images or scenes as continuous functions of coordinates and are often paired with modern encodings~\cite{dupont2021coin, dupont2022coin++}, including Fourier features~\cite{tancik2020fourier}, periodic activations~\cite{sitzmann2020implicit}, adaptive coordinates~\cite{martel2021acorn}, vector-quantized auto-decoders~\cite{takikawa2022variable}, and multiresolution hash grids such as Instant-NGP~\cite{mueller2022instant}. They are also widely used as implicit priors for image reconstruction and compression~\cite{ulyanov2018deep,dupont2021coin}. However, their implicit parameterization typically obscures explicit spatial ownership, making direct editing, localized reallocation of capacity, and budget-aware compression less straightforward. \name instead targets an explicit primitive-based structure with localized responsibilities and predictable query cost, while remaining fully differentiable. {Recent discontinuity-aware neural fields are closer to mesh-based neural representations than to point- or splat-based image primitives. DANF~\cite{belhe2023discontinuity} assumes input discontinuity curves, constructs a curved triangulation constrained by them, stores features on mesh vertices and discontinuous edges, and decodes discontinuity-aware interpolated features with a shallow MLP. NFLD~\cite{liu20252d} further learns unknown discontinuities on a triangle mesh by defining local discontinuous feature functions over vertex one-rings, treating all mesh edges as potentially discontinuous, and jointly optimizing discontinuity magnitudes with the field on a mesh initialized from Canny edges~\cite{canny2009computational} followed by TriWild triangulation~\cite{Hu:2019:TRT:3306346.3323011}. Earlier hybrid random-access texture representations also combine compact decoding with learned predictors~\cite{song2015vector}, but do not provide the explicit ownership and adjacency structure of \name. In contrast, \name is a compact, explicit primitive-based image representation that captures sharp, content-aligned discontinuities without requiring prescribed curves or edge-to-mesh preprocessing, while enabling adaptive budget control and efficient fitting through fixed-size, per-pixel top-$K$ evaluation.} {Earlier classical graphics has also explored image and texture representations with embedded discontinuities, including {Scale-Dependent Reproduction of Pen-and-Ink Illustrations}~\cite{salisbury1996scale}, {Feature-Based Textures}~\cite{ramanarayanan2004feature}, {Bixels}~\cite{tumblin2004bixels}, Pinchmaps~\cite{tarini2005pinchmaps}, and Real-Time Rendering of Textures with Feature Curves~\cite{parilov2008real}. These methods explicitly encode sharp boundaries or feature curves for magnification and filtering, whereas \name targets a differentiable, per-instance optimized image representation with adaptive primitives and learnable soft-to-sharp ownership.}

\paragraph{Point-based and splat-based rendering.}
Point-based rendering~\cite{pfister2000surfels, botsch2003high, zwicker2001surface, botsch2005high} has a long history (e.g., surfels~\cite{pfister2000surfels} and surface splatting~\cite{zwicker2001surface}),
with anisotropic filtering and efficient evaluation. More recently, 2D Gaussian Splatting~\cite{huang20242d} (2D version of 3D Gaussian Splatting~\cite{kerbl20233d}) has been adapted for compact image representations, including
Image-GS~\cite{imagegs2025} and GaussianImage~\cite{gaussianimage2024}.
In contrast, \name replaces kernel overlap with a temperature-controlled soft partition of unity induced by distance-based scores, which yields clearer spatial ownership, explicit adjacency, and more direct signals for budget control (e.g., pruning and densification). 

Point-based methods are closely related to Voronoi diagrams and their variants (as well as their dual graphs, e.g., Delaunay Triangulation), which have been extensively explored in related areas~\cite{di2025spherical, govindarajan2025radiant, gu2024tetrahedron}. We next summarize the necessary preliminaries and review the relevant prior work.

\section{Preliminaries}
\label{sec:background_voronoi_power}
\paragraph{Voronoi, Apollonius, and Power Diagrams}

\begin{figure}[t]
  \centering
  \begin{overpic}[width=0.95\linewidth]{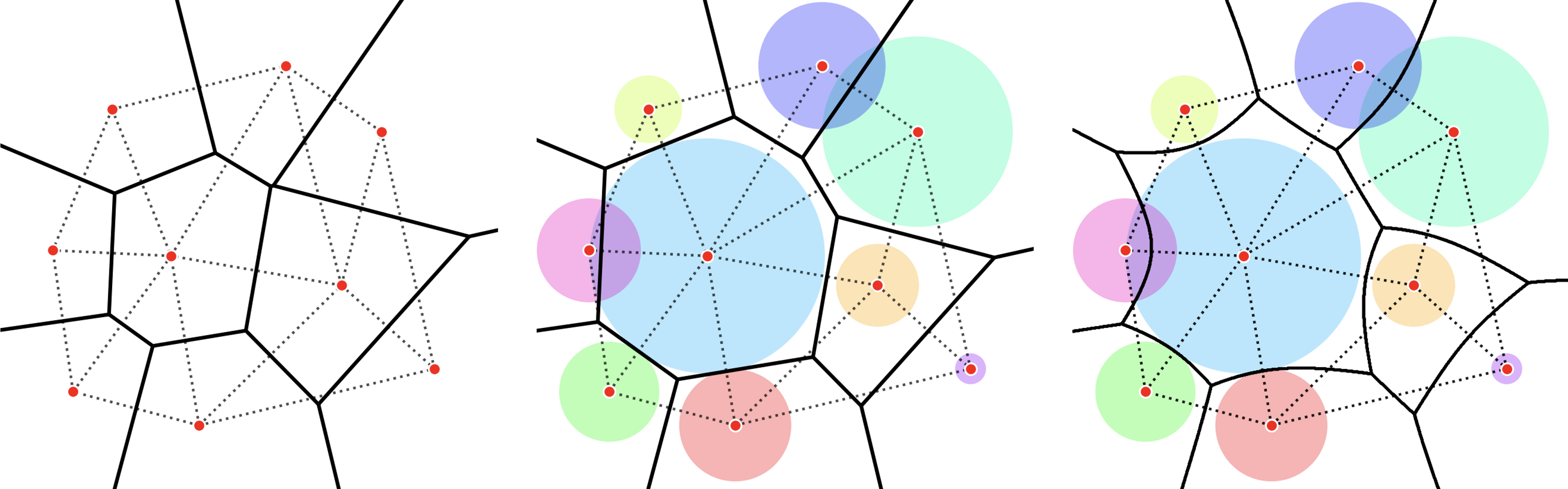}
    \put(4,-4.5){\small Voronoi diagram}
    \put(40,-4.5){\small Power diagram}
    \put(71,-4.5){\small Apollonius diagram}
  \end{overpic}
\vspace{2mm}
\caption{Voronoi, power (Laguerre), and additively weighted Voronoi (Apollonius) diagrams in 2D. Red points denote sites. Solid black segments show the induced partition boundaries. Dashed gray connections visualize site adjacencies (the dual graph). Colored disks illustrate per-site weights/radii for the weighted variants.}
  \label{fig:vd_pd_ad}
  \vspace{-2mm}
\end{figure}

Let $\Omega\subset\mathbb{R}^d$ denote a bounded domain and let
$\mathcal{P}=\{p_i\in\Omega\}_{i=1}^N$ be a set of sites (generators).
{We briefly review the Euclidean, weighted, and anisotropic constructions most relevant to our method.}

\paragraph{Voronoi diagram.}
The (Euclidean) Voronoi diagram~\cite{voronoi1908nouvelles,aurenhammer1991voronoi}
partitions $\Omega$ into cells
\begin{equation}
\Omega_i^{\mathrm{vor}}
=
\Big\{x\in\Omega~\big|~
\|x-p_i\|\le \|x-p_j\|,\ \forall j\neq i
\Big\}.
\label{eq:vor_cell}
\end{equation}
Each cell contains points closest to its generator in Euclidean distance, yielding
piecewise-linear boundaries (in 2D, polygonal cells). Voronoi diagrams provide
explicit ownership and adjacency (via shared cell boundaries), properties that are
useful for structured representations.

\paragraph{Additively weighted Voronoi (Apollonius) diagrams.}
Additively weighted Voronoi diagrams compare distances after subtracting a
per-site radius (equivalently, distances to weighted balls), yielding Apollonius
diagrams~\cite{emiris2006predicates}. Defining the additively weighted distance
\begin{equation}
d_i^{\mathrm{apo}}(x)=\|x-p_i\|-r_i,
\label{eq:apollonius_dist}
\end{equation}
the Apollonius (additively weighted Voronoi) cell is
\begin{equation}
\Omega_i^{\mathrm{apo}}
=
\Big\{x\in\Omega~\big|~
d_i^{\mathrm{apo}}(x)\le d_j^{\mathrm{apo}}(x),\ \forall j\neq i
\Big\}.
\label{eq:apollonius_cell}
\end{equation}
Intuitively, larger $r_i$ expands the influence of site $i$ by shifting the
distance comparison, providing an intuitively interpretable ``influence radius''
for each site.

\paragraph{Power diagrams / Laguerre tessellations.}
Power diagrams (also known as Laguerre diagrams) generalize Voronoi diagrams by
adding weights to the squared distance~\cite{aurenhammer1987power}. Given
weights $\{w_i\}$, the power distance is
\begin{equation}
d_i^{\mathrm{pow}}(x)=\|x-p_i\|^2-w_i,
\label{eq:power_dist}
\end{equation}
and the corresponding cell is
\begin{equation}
\Omega_i^{\mathrm{pow}}
=
\Big\{x\in\Omega~\big|~
d_i^{\mathrm{pow}}(x)\le d_j^{\mathrm{pow}}(x),\ \forall j\neq i
\Big\}.
\label{eq:power_cell}
\end{equation}
{Unlike the Apollonius construction, the comparison in Eq.~\eqref{eq:power_dist}
is quadratic in $x$. Power diagrams are also dual to regular triangulations
(weighted Delaunay).}

\paragraph{Anisotropic Voronoi and centroidal tessellations.}
Voronoi-like partitions can be extended to anisotropic settings by replacing the
Euclidean norm with a metric. A common form uses an SPD matrix
$G_i\succ 0$ to define a site-dependent norm
\begin{equation}
\|x-p_i\|_{G_i}=\sqrt{(x-p_i)^\top G_i (x-p_i)}.
\label{eq:anis_norm}
\end{equation}
Anisotropic and centroidal variants have been studied for mesh generation and
approximation under such metrics. {Figure~\ref{fig:vd_pd_ad} illustrates the Euclidean Voronoi, power/Laguerre, and Apollonius variants
together with their induced adjacency structures.}

\paragraph{Connection to \name}
{\name adopts an additively weighted formulation under an anisotropic norm.}
Concretely, each site defines an SPD metric $G_i$ and an additive radius $r_i$,
and we compare sites using
\begin{equation}
d_i(x)=\|x-p_i\|_{G_i}-r_i,
\label{eq:our_score_hard}
\end{equation}
{which yields an anisotropic Apollonius-style comparison rather than a
power/Laguerre (quadratic) score. Instead of forming a hard partition via
$\arg\min_i d_i(x)$, \name relaxes this construction into a differentiable soft
partition of unity through a temperature-controlled softmax~\cite{di2025spherical},
enabling stable gradients for optimization while still allowing sharp,
content-aligned transitions.}

\paragraph{Differentiable Voronoi representation and its applications.}
Voronoi diagrams and their variants have been extensively used in classical geometric processing and modeling, including shape reconstruction and generation~\cite{boltcheva2017surface,amenta1998surface,yan2013efficient,amenta2001power, xu2023globally, guo2024medial, amenta1998new}, shape analysis~\cite{amenta2001power,dou2022coverage,wang2024coverage,li2015q, wang2022computing, wang2024mattopo}, and motion planning~\cite{takahashi1989motion,bhattacharya2007voronoi}. Beyond classical computational geometry, several recent works make Voronoi-style partitions differentiable and optimize generators as learnable primitives. VoronoiNet~\cite{williams2020voronoinet} uses soft Voronoi cells with local support as general functional approximators. In geometry, VoroMesh~\cite{maruani2023voromesh} and VoroLight~\cite{lu2025vorolight} learn Voronoi-based surface and volumetric structures by optimizing generator locations. Differentiable Voronoi diagrams have also been used for design and optimization, including cellular topology optimization~\cite{feng2023cellular} and free-form floor-plan generation~\cite{wu2024free}. {Related structured differentiable representations also extend beyond the 2D image plane: Differentiable Surface Triangulation~\cite{rakotosaona2021differentiable} optimizes soft triangulations, Spherical Voronoi~\cite{di2025spherical} introduces a differentiable temperature-controlled Voronoi-like partition on the unit sphere for modeling view-dependent appearance, and RadiantFoam~\cite{govindarajan2025radiant} uses volumetric-mesh primitives for real-time differentiable ray tracing.} Unlike these methods, \name focuses on per-image 2D signal representation, using an anisotropic, additively weighted (Apollonius) score with a learnable per-site temperature. Leveraging locality, it enables GPU-resident, constant-cost evaluation through a per-pixel top-$K$ propagation scheme with jump-flooding-inspired updates~\cite{rong2006jumpflooding}, delivering efficient image fitting while preserving explicit ownership and adjacency.

\section{Method}
\label{sec:method}

\begin{figure*}
\centering
\includegraphics[width=\textwidth]{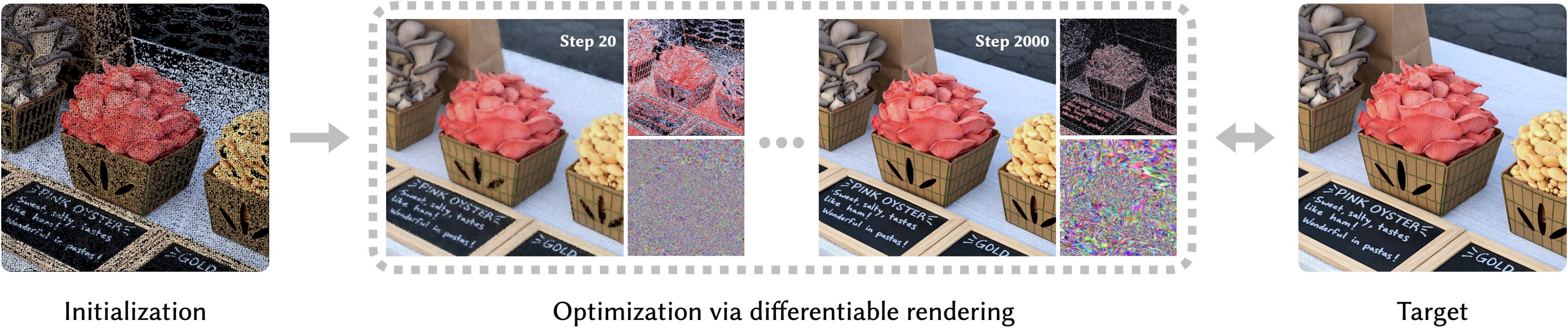}
\vspace{-5mm}
\caption{The Pipeline of \name. \textbf{Left:} Given an input image, we initialize sites using content-adaptive sampling that places more sites in high-gradient regions (edges and textures).
\textbf{Middle:} We jointly optimize site parameters including positions, colors, radii, and blending temperatures ($\tau$). The \textit{Voronoi Partition} row (bottom right) shows the evolving spatial decomposition, while the \textit{Tau Heatmap} (top right) visualizes the learned blending sharpness: warm colors indicate \emph{harder}, sharper transitions (high $\tau$) suited for edges, while cool colors indicate \emph{softer} transitions (low $\tau$) suited for soft gradients. Through iterative densification and pruning, the site count reduces from $128$K to $25$K while preserving quality.
\textbf{Right:} The final output includes the reconstructed image, Apollonius partition boundaries, and optimized site distribution. We also visualize the content-adaptive site density and learned partition structure.}
\label{fig:main-diagram}
\end{figure*}

{We model an image as a set of $N$ anisotropic sites with colors and learnable temperatures. Each site defines an additively weighted score, and the image is rendered by a soft partition of unity over a small top-$K$ site set. The resulting representation is explicit and spatially local, but unlike a classical hard Voronoi partition, it remains differentiable because multiple nearby sites can contribute at once. We next describe the site parameterization, how we maintain the per-pixel top-$K$ list with Jump Flood warm-up and subsequent single-pass local refreshes, and how we optimize and adapt the site budget during training. An overview of the full pipeline is shown in Figure~\ref{fig:main-diagram}.}

\subsection{Site Representation}
We represent an image with $N$ sites. Each site $i$ has:
\begin{itemize}[leftmargin=1.2em]
  \item position $p_i \in \mathbb{R}^2$
  \item temperature parameter $\log\tau_i$ ($\tau_i = \exp(\log\tau_i)$)
  \item radius (additive weight) $r_i$
  \item color $c_i \in \mathbb{R}^3$
  \item anisotropy direction $u_i \in \mathbb{R}^2$, $\|u_i\|=1$
  \item log-anisotropy $a_i$ controlling aspect with $\det(G_i)=1$
\end{itemize}

During optimization we clamp positions to image bounds and re-normalize
$u_i$ after each step.

{We use the following notation throughout. Site $i$ has position $p_i$, color $c_i$, radius $r_i$, temperature $\tau_i$, anisotropy direction $u_i$, and anisotropy scalar $a_i$. The matrix $G_i$ is the corresponding SPD metric, so $a_i$ and $G_i$ are linked: $a_i$ sets the eigenvalues $e^{a_i}$ and $e^{-a_i}$ of $G_i$ along $u_i$ and its orthogonal direction. For a query pixel $x$, $\mathcal{C}(x)$ denotes the maintained top-$K$ candidate set and $c(x)$ the rendered color. We use $K$ for candidate list size, $F_d$ and $F_p$ for densify/prune frequencies, and $\alpha,\varepsilon$ for the densify-score constants.}

We define an anisotropic metric per site. Let $v_i$ be a unit vector orthogonal to $u_i$:
\begin{equation}
    G_i = e^{a_i} u_i u_i^\top + e^{-a_i} v_i v_i^\top .
\end{equation}
This SPD (symmetric positive definite) metric has $\det(G_i)=1$, so it changes aspect without area scaling.

We use the norm induced by $G_i$,
\[
\|x-p_i\|_{G_i} = \sqrt{(x-p_i)^\top G_i (x-p_i)}.
\]
To make the model resolution-invariant, let $s = 1/\max(H,W)$. We define a signed,
normalized additively weighted distance score
\begin{equation}
d_{\text{mix}}(x,i) = \|x-p_i\|_{G_i}\,s - r_i\,s,
\end{equation}

where any constant scale can be absorbed into $\tau_i$, so we omit a separate
logits factor. 

{Intuitively, $r_i$ controls how far site $i$ reaches before its score becomes unfavorable, $G_i$ controls directional stretch and orientation, and $\tau_i$ controls how sharply that score is converted into mixture weights. Larger radii increase effective support, larger $|a_i|$ elongates the footprint, and larger $\tau_i$ sharpens transitions.}

\subsection{Soft Additively Weighted Voronoi Rendering}
Given a pixel $x$, each site produces a logit:

\begin{equation}
\ell_i(x) = -\tau_i\, d_{\text{mix}}(x,i).
\end{equation}

Each site has its own temperature $\tau_i$.

We compute weights over a top-$K$ site set $\mathcal{C}(x)$:
\begin{equation}
w_i(x) = \frac{\exp(\ell_i(x))}{\sum_{j\in\mathcal{C}(x)}\exp(\ell_j(x))}.
\end{equation}

The pixel color is:
\begin{equation}
c(x) = \sum_{i\in\mathcal{C}(x)} w_i(x)\,c_i.
\end{equation}

{We use the term \emph{diagram} in the geometric sense that the score field induces an Apollonius-Diagram-style neighborhood structure. The rendered model itself is not a strict hard nearest-site partition: the softmax produces a soft partition of unity over $\mathcal{C}(x)$, so several nearby sites may jointly explain a pixel. Restricting the blend to top-$K$ is both a computational device and a locality prior, since each pixel depends only on a small explicit neighborhood of candidate sites.}

\begin{figure}
    \centering
    \includegraphics[width=1.0\linewidth]{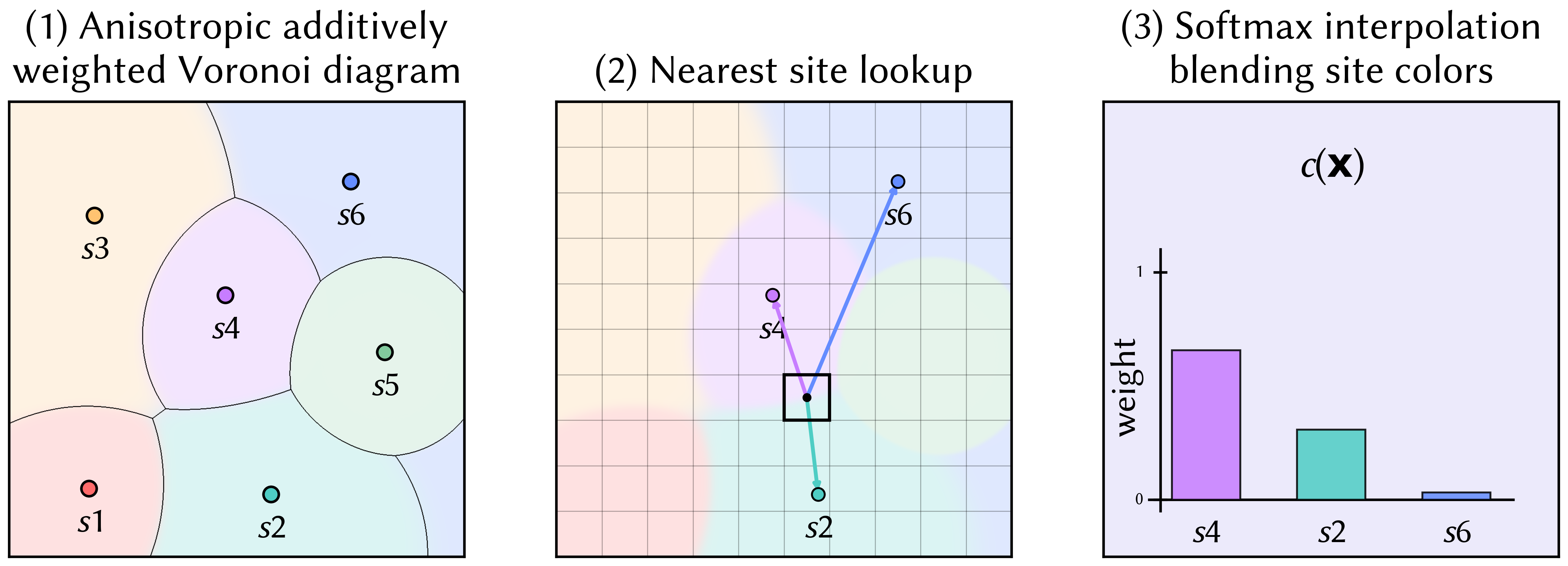}
    \caption{\textbf{Representation diagram of our soft anisotropic additively weighted rendering.} (1) Sites induce an anisotropic additively weighted diagram under the score $d_{\text{mix}}(x,i)= s\lVert x-p_i\rVert_{G_i}-s\,r_i$. Here $p_i$ sets position, $r_i$ controls reach, and $G_i$ controls orientation and anisotropic stretch. (2) For a query pixel $x$, we consider a small set of nearest candidate sites $\mathcal{C}(x)$ under this score. (3) We compute temperature-controlled softmax weights $w_i(x)\propto \exp\!\big(-\tau_i d_{\text{mix}}(x,i)\big)$ over $i\in\mathcal{C}(x)$ and blend site colors to render $c(x)=\sum_{i\in\mathcal{C}(x)} w_i(x)\,c_i$, where larger $\tau_i$ yields sharper boundaries and smaller $\tau_i$ yields softer blending. }
    \label{fig:representation}
\end{figure}

{Our score adopts an anisotropic additively weighted form, $d_{\text{mix}}(x,i)= s\lVert x-p_i\rVert_{G_i}-s\,r_i$ with $s=1/\max(H,W)$, which can be interpreted as a signed distance to an oriented anisotropic ball with effective radius $r_i$. Compared to power/Laguerre distances, the square-root form keeps $r_i$ more geometrically interpretable under anisotropy and better decouples \emph{reach} ($r_i$) from \emph{hardness} ($\tau_i$). The temperature-controlled softmax $w_i(x)\propto\exp(-\tau_i d_{\text{mix}}(x,i))$ then provides a continuous soft-to-hard knob, so optimization can start with smooth responsibilities and progressively sharpen transitions where needed. This behavior is analyzed in Section~\ref{sec:ablation}. Figure~\ref{fig:representation} illustrates the rendering process and the role of the soft partition.}

\subsection{Top-$K$ Propagation}
\label{sec:topk-propagation}
{Evaluating all sites per pixel is expensive. We therefore maintain a small per-pixel top-$K$ site set that approximates the top-$K$ sites under the same additively weighted score used for rendering. This can be viewed as an approximate $K$-th order diagram query under the SAD score, but the practical goal is locality: each pixel tracks only a small explicit neighborhood of competitive sites.}

{We use Jump Flooding Algorithm (JFA)~\cite{rong2006jumpflooding} to warm up the candidate field. Standard JFA propagates seed information across the image with a coarse-to-fine schedule $B/2,B/4,\ldots,1$. Let $B = 2^{\lceil \log_2 \max(H,W) \rceil}$. At candidate-refresh event $t$, the jump step is $s_t = \max\!\left(1, B / 2^{\min(t,\log_2 B - 1)+1}\right)$, so the step sizes are $B/2, B/4, \ldots, 1$ and then remain 1. Thus JFA serves as the early long-range warm-up; once the jump step reaches 1, later training refreshes are single-pass immediate-neighbor updates.}

{Temporal reuse warm-starts the update from the previous top-$K$ field. This is effective because site parameters usually change gradually, so most strong candidates remain competitive from one refresh to the next. Spatial propagation then merges candidates from a fixed local neighborhood (self + 4 neighbors) at the current step. A small nonlocal probe set is still needed because purely local propagation can miss a newly competitive distant site, especially after densification, pruning, or larger site motion. These occasional nonlocal candidates seed regions that can then be propagated locally in later refreshes, improving recovery speed and practical convergence of the maintained top-$K$ field.}
We formalize the update as:
\begin{equation}
\tilde{\mathcal{C}}_t(x)=\mathcal{C}_{t-1}(x)\cup \mathcal{P}_t(x)\cup \mathcal{G}_t(x),
\qquad
\mathcal{C}_t(x)=\operatorname{topK}_{i\in\tilde{\mathcal{C}}_t(x)} \ell_i(x),
\end{equation}

{where $\mathcal{P}_t(x)$ denotes sites propagated from the self + 4-neighbor stencil and $\mathcal{G}_t(x)$ denotes the small global probe set.}

\begin{figure}
    \centering
    \includegraphics[width=1.0\linewidth]{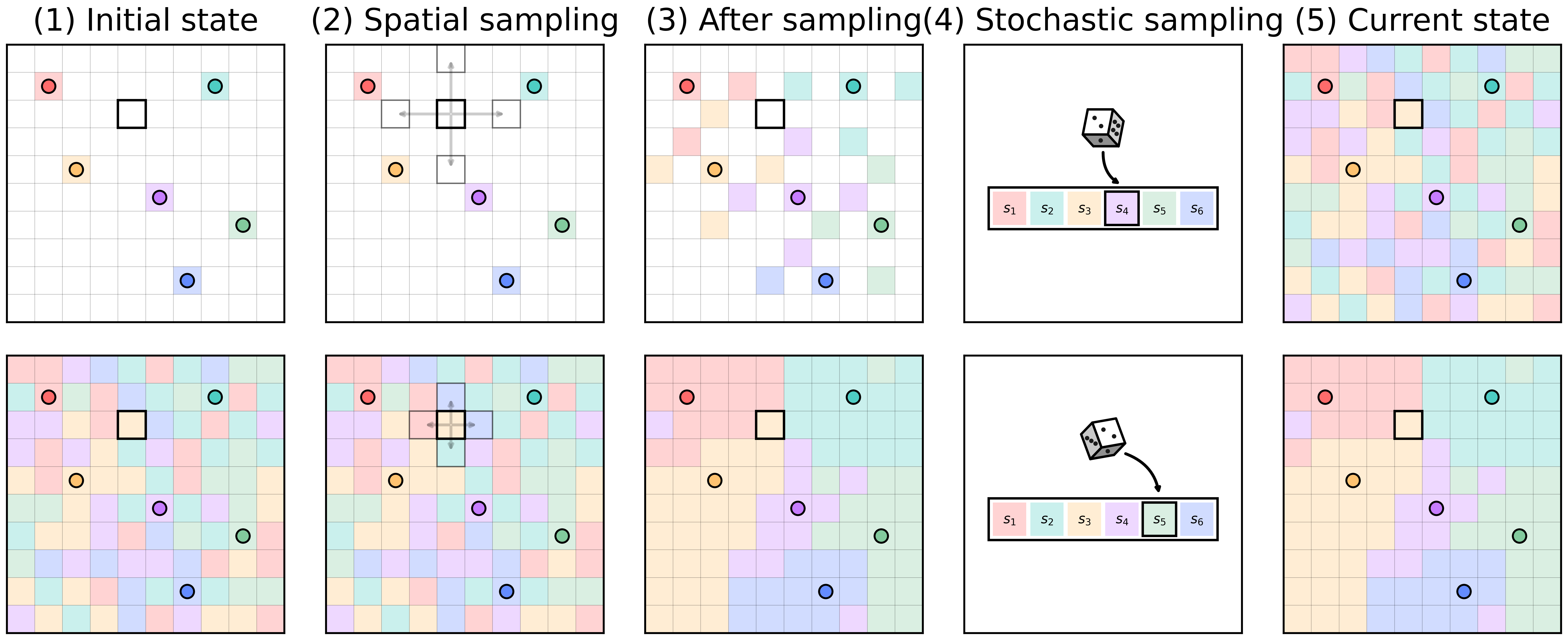}
    \caption{\textbf{Top-$K$ propagation algorithm for maintaining per-pixel candidate site sets at constant cost.} (1) Initialize each pixel’s list with the previous top-$K$ set $\mathcal{C}_{t-1}(x)$ (temporal reuse). (2) Spatial propagation: merge candidate site IDs from a fixed neighborhood (self + 4 neighbors) using the current jump step $s_t$. (3) After propagation, each pixel holds an expanded candidate pool. (4) A small global probe step adds nonlocal candidate IDs when spatial propagation alone may miss distant competitors. (5) Score all candidates and keep the top $K$ by $\ell_i(x)$ to obtain the updated set $\mathcal{C}_t(x)$ (an approximate $K$-th order diagram update under the SAD score).}
    \label{fig:topk_propagation}
\end{figure}

The parallel nature of the algorithm makes it a perfect fit for modern GPUs. Each thread has nearly constant work and coalesced memory access, resulting in high utilization and cost efficiency. Although a larger site count might require more iterations, by design, the computational cost per pass stays constant, which is an essential property for fast encoding. 
Figure~\ref{fig:topk_propagation} summarizes the update steps.

\subsection{Initialization}
We use a content-adaptive, gradient-weighted initialization similar in spirit to
Image-GS. Positions are sampled at pixel centers from a mixture of normalized
gradient magnitude and a uniform prior:
\begin{equation}
P_{\text{init}}(x) =
(1-\lambda_{\text{init}})\,
\frac{\|\nabla I(x)\|_2}{\sum_{x'} \|\nabla I(x')\|_2}
\;+\;
\lambda_{\text{init}}\frac{1}{H\cdot W},
\qquad \lambda_{\text{init}}\in[0,1].
\end{equation}
The gradient term concentrates sites in high-frequency regions, while the
uniform term preserves coverage of soft areas. Each site is initialized with
the target pixel color at its sampled location, and other parameters use fixed
defaults. This simple initialization accelerates convergence but is not a novel
contribution.

\subsection{Adaptive Budget: Densify and Prune}
We adapt the number of active sites during training. Define soft responsibilities:
\begin{equation}
m_i = \sum_x w_i(x),\qquad
E_i = \sum_x w_i(x)\,\|c(x)-I(x)\|^2.
\end{equation}
We use an error-density heuristic:
\begin{equation}
s_i = \frac{E_i}{\max(m_i,\varepsilon)^\alpha}.
\end{equation}
{This score favors sites whose residual error remains high relative to the soft mass they currently explain. As a result, even a lower-responsibility site can be selected for splitting if it sits in an underserved region with concentrated error, which is the regime where allocating extra local capacity is most useful.}
We set $\varepsilon=10^{-8}$ and ignore sites with $m_i \le 1$ in densify scoring.
We densify every $F_d$ iterations (densify frequency) within a window
$[t_d^{\text{start}}, t_d^{\text{end}}]$
by splitting the top percentile of sites by $s_i$ (ignoring very small-mass
sites). For each selected site, we estimate a residual-weighted
centroid and covariance from per-site statistics and split along the principal
axis; if statistics are insufficient, we fall back to the local image gradient.
Children are offset along this axis by $0.5\sqrt{m_i}$ pixels (clamped to
[1.5, 48]), inherit the parent parameters, and apply $\log\tau \leftarrow \log\tau-0.25$,
$r \leftarrow 0.85\,r$, and $a$ from the covariance (clamped to $[-2,2]$; or
$0.8\,a$ when statistics are unavailable). Colors are re-sampled from the
target at the new positions. We prune every $F_p$ iterations (prune frequency) within
$[t_p^{\text{start}}, t_p^{\text{end}})$ by removing the bottom percentile of
sites under the removal-delta score (normalized by the number of valid pixels),
optionally delaying pruning until densification ends.
When a target BPP is specified, we scale the base densify/prune percentiles with
a schedule simulator to match the expected final site count under the iteration
budget, via a 1D search over a shared scale factor.

\paragraph{Removal delta (per-site prune signal).}
Let $\hat c=c(x)$ be the rendered color at pixel $x$. Given a pixel with weights $w$ that sum to 1, removing site $k$ and renormalizing the remaining weights yields
\begin{equation}
\hat c' = \frac{\hat c - w_k c_k}{1-w_k}.
\end{equation}
{The local loss increase $\Delta\mathcal{L}(x) = \|\hat c'-I(x)\|^2-\|\hat c-I(x)\|^2$ can be computed in closed form from $\hat c$, $I(x)$, $c_k$, and $w_k$, and accumulated per site to form an efficient prune score. Intuitively, this removal delta estimates how much the reconstruction would deteriorate if a site were deleted and the remaining contributors were renormalized, so sites with small accumulated delta are natural prune candidates.}

Implementation details, including top-$K$ list storage and gradient accumulation, are
described in the Implementation section.

\subsection{Top-$K$ Site Set Storage and Updates}
{We store the per-pixel top-$K$ site set as a fixed-size list ($K=8$), packed into two textures for coalesced access. The update algorithm itself is described in Section~\ref{sec:topk-propagation}; here we focus on the storage layout and constant-time maintenance of the packed top-$K$ list. Each refresh operates directly on this packed fixed-size layout and maintains the list with insertion into a fixed-size array, yielding O($K$) time per insert (constant in practice for small $K$). For full refreshes, we can run multiple passes and optionally reseed with JFA when the top-$K$ list is invalidated.}

\subsection{Cost Profile and Bandwidth Tradeoff}
{Our per-pixel computation is constant and does not scale with the total number of sites. Maintaining the per-pixel top-$K$ field costs O($P\!\cdot\!K$), and rendering also costs O($P\!\cdot\!K$), both with small constants because each pixel only evaluates a fixed-size candidate list. By contrast, Image-GS rasterizes every Gaussian overlapping each tile, incurring O($P\!\cdot\!G$) work, where $G$ is the average number of overlaps per tile (top-$K$ normalization does not reduce this scan). It also rebuilds tile bins every render via global intersection generation and sorting, adding an O($N_{\mathrm{int}}\log N_{\mathrm{int}}$) term, where $N_{\mathrm{int}}$ is the total number of intersections.}

These algorithmic differences translate into a computation and bandwidth tradeoff on
GPU: our kernels are bandwidth-bound but highly regular (packed 16-byte
quantized site records in the inference/candidate pipelines, coalesced top-$K$
reads, and shared-memory reductions), while
Gaussian methods incur both higher compute and substantial global memory traffic
from sorting, scatter or gather, and atomic-heavy backprop. This fixed-cost
structure explains the large practical speedups we observe even at high
resolutions.

\subsection{Gradient Accumulation}
\begin{figure}[t]
  \centering
  \includegraphics[width=\linewidth]{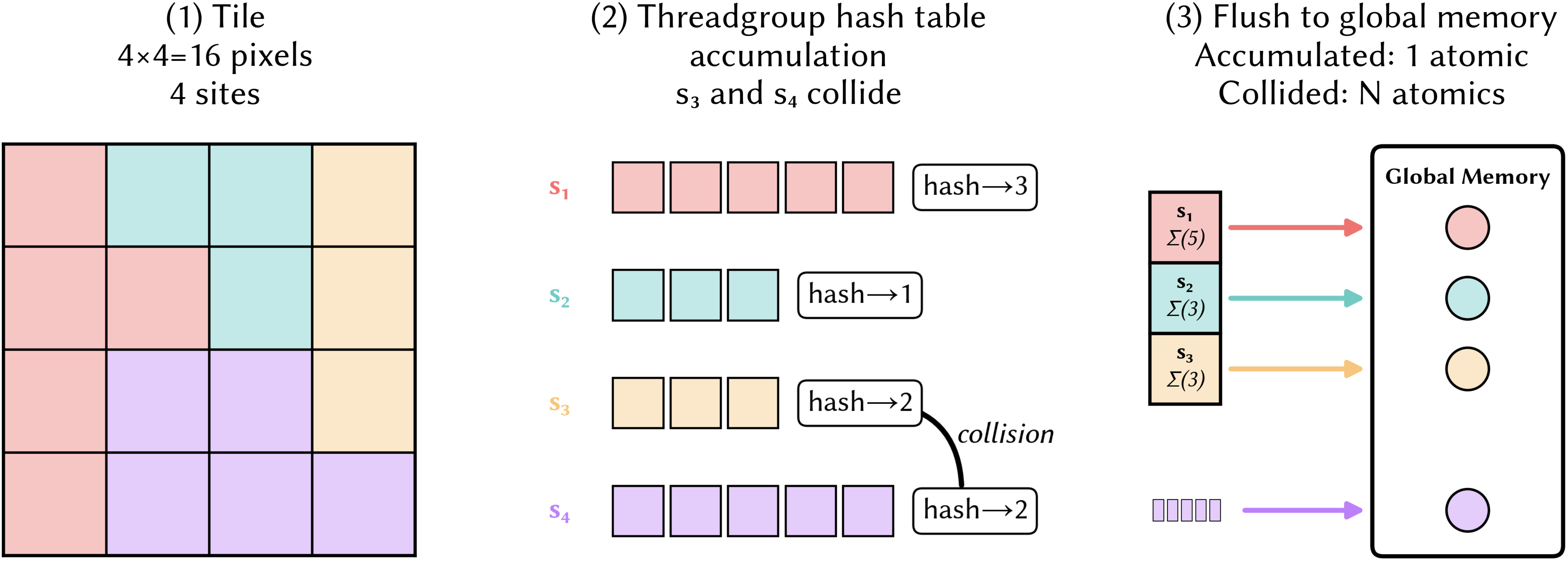}
  \caption{\textbf{Threadgroup hash table gradient accumulation.}
  (1) Each tile processes multiple pixels, with each pixel potentially affected by multiple sites (shown in different colors).
  (2) Sites are hashed into a threadgroup-local hash table where gradients are accumulated using local atomics. When sites collide (e.g., $s_3$ and $s_4$ both hash to slot 2), open addressing is used with bounded probing.
  (3) After all pixels in the tile are processed, accumulated gradients are flushed to global memory. Sites that were successfully accumulated in the hash table require only 1 atomic write, while overflowed or collided sites may require multiple atomic writes. This approach significantly reduces global atomic contention compared to naive per-pixel atomics.}
  \label{fig:hash_gradient_accumulation}
\end{figure}

Naive per-pixel atomic accumulation suffers from high contention because many
pixels update the same sites, creating heavy contention on global atomics,
serialized updates, and scattered writes that thrash caches. To reduce
contention while keeping the pipeline fully GPU-driven, we use a tiled
threadgroup hash reduction (Algorithm~\ref{alg:hash_grad} and
Figure~\ref{fig:hash_gradient_accumulation}).

\begin{algorithm}[t]
  \caption{Threadgroup hash reduction for gradient accumulation.}
  \label{alg:hash_grad}
  \begin{algorithmic}[1]
    \State Initialize shared hash table keyed by siteID.
    \For{each pixel in tile}
      \For{each top-$K$ site for pixel}
        \State Probe insert key.
        \State Accumulate gradients in shared memory.
      \EndFor
    \EndFor
    \State Synchronize threads.
    \For{each entry in hash table}
      \State Flush to global buffers (one atomic per site per tile).
    \EndFor
    \If{table overflows}
      \State Fallback: accumulate those sites via global atomics.
    \EndIf
  \end{algorithmic}
\end{algorithm}

This replaces $O(P\!\cdot\!K)$ scattered global atomics with a small number
of localized reductions per tile, improving cache locality and reducing
contention.

\section{Implementation Details}
We implement the full training and rendering pipeline as a single GPU-resident
renderer using compute shaders, with backends for Metal, CUDA, and WebGPU. 
All forward kernels and their corresponding gradients are hand-derived and 
implemented directly. We do not rely on automatic differentiation frameworks. 
The entire pipeline, including forward rendering, backward differentiation, candidate propagation, and Adam optimization, executes without any CPU work 
during training iterations, eliminating host-device roundtrips that typically 
dominate latency in differentiable rendering systems. We validate the method on
a broad range of devices, including multiple Apple Silicon GPUs (Metal)
and NVIDIA GPUs (CUDA), and keep a shared set of default parameters across
backends to avoid device-specific tuning. 
Optimization uses Adam with per-parameter base learning rates and fixed
betas/epsilon, and the densification/pruning and candidate-update schedules
follow a single shared default configuration across backends.
The implementation details and the full list of runtime parameters/constants are summarized in the Appendix.

\section{Evaluation}
\label{sec:evaluation}
\subsection{Experimental Setup}
We evaluate reconstruction quality using three complementary metrics: PSNR (peak signal-to-noise ratio, higher is better), SSIM (structural similarity index, higher is better), and LPIPS (learned perceptual image patch similarity, lower is better). We also report training time and rendering performance.

\paragraph{Our method (\name)}
We use $K=8$ top-$K$ sites per pixel and train for 4000 iterations with adaptive pruning.
We optimize site parameters with Adam using a pure MSE reconstruction loss:
\[
\mathcal{L} = \mathbb{E}_x\, \|c(x) - I(x)\|^2,
\]
and clamp parameters to valid ranges (positions in image bounds, $\log\tau \in [2,20]$,
$r \in [1,512]$, $a \in [-2,2]$, colors in $[0,1]$), with $u_i$ re-normalized
after each step. Notably, we do \emph{not} use SSIM loss during training, relying
entirely on pixel-level MSE.
Top-$K$ propagation uses $J{=}4$ injected site IDs per pass. The base densify/prune schedule uses
densify every 20 iterations from 20--3000 with percentile 0.01 and
$\alpha{=}0.7$, and prune every 40 iterations from 100--3000 with percentile
0.033; when pruning during densification is disabled, pruning starts after
densification ends. When a target BPP is specified, we scale the base
densify/prune percentiles with a schedule simulator that matches the expected
final site count under the iteration budget (by scaling both percentiles with
a single factor, capped at 0.95). Densify scores use $\varepsilon{=}10^{-8}$ and
ignore sites with $m_i \le 1$.

\paragraph{Baselines.}
We compare against the SOTA methods Instant-NGP~\cite{mueller2022instant} and Image-GS~\cite{imagegs2025} using their official implementations.

Note that both \name and Image-GS can be viewed as normalized mixtures of exponentiated distance scores over a restricted top-$K$ set. The difference is in parameterization and the geometric interpretation: \name uses an additively weighted anisotropic distance score with a det-1 metric and a separate temperature, while Image-GS uses Gaussian covariance and kernel overlap. Table~\ref{tab:structure_matrix} summarizes how \name compares to common 2D representations across structural properties.

\providecommand{\cmark}{\ensuremath{\checkmark}} \providecommand{\xmark}{\ensuremath{\times}} \providecommand{\pmark}{\ensuremath{\triangle}} \newcommand{\hdr}[1]{\parbox[c]{0.12\linewidth}{\centering #1}} 

\begin{table*}
\centering
\footnotesize
\renewcommand{\arraystretch}{1.12}
\setlength{\tabcolsep}{0pt} 

\caption{Qualitative comparison of 2D image representations by structural properties.}
\label{tab:structure_matrix}
\vspace{-3mm}
\resizebox{\textwidth}{!}{
\begin{tabular}{C{0.36\linewidth} C{0.12\linewidth} C{0.14\linewidth} C{0.08\linewidth} C{0.21\linewidth} C{0.19\linewidth} C{0.19\linewidth}}
\toprule
\textbf{Method} &
\shortstack{\textbf{Explicit}\\[-1pt]\textbf{primitives}} &
\shortstack{\textbf{Convex}\\[-1pt]\textbf{output}} &
\textbf{Local} &
\shortstack{\textbf{Handles}\\[-1pt]\textbf{discontinuities}} &
\shortstack{\textbf{Const.}\\[-1pt]\textbf{query cost}} &
\shortstack{\textbf{No edge/mesh}\\[-1pt]\textbf{preproc.}}\\
\midrule
\textbf{Gaussian splats~\cite{gaussianimage2024,imagegs2025}} 
& \cmark & \pmark & \pmark & \pmark & \xmark & \cmark \\

\textbf{Neural Fields~\cite{dupont2021coin,mueller2022instant}} 
& \xmark & \xmark & \xmark & \pmark & \cmark & \cmark \\

\textbf{Discontinuity-aware Neural Fields~\cite{belhe2023discontinuity,liu20252d}}$^\dagger$
& \pmark & \xmark & \cmark & \cmark & \pmark & \xmark \\

\textbf{RBF~\cite{buhmann2000rbf}} 
& \cmark & \xmark & \pmark & \pmark & \pmark & \cmark \\

\textbf{B-spline~\cite{deboor1978splines}} 
& \cmark & \cmark & \cmark & \pmark & \cmark & \cmark \\

\midrule
\textbf{\name (ours)} 
& \cmark & \cmark & \cmark & \cmark & \cmark & \cmark \\
\bottomrule
\end{tabular}
}

\smallskip
\begin{minipage}{0.98\textwidth}
\footnotesize
Legend: \cmark\ = yes, \pmark\ = partial/depends, \xmark\ = no.
\end{minipage}
\end{table*}

\begin{table}[t]
  \centering
  \small
  \resizebox{0.98\linewidth}{!}{
  \begin{tabular}{p{0.28\linewidth} p{0.34\linewidth} p{0.34\linewidth}}
    \toprule
    \textbf{Property} & \textbf{\name} & \textbf{Image-GS} \\
    \midrule
    Primitive & Anisotropic additively weighted Voronoi cell & Anisotropic 2D Gaussians \\
    Logit/weight form & $-\tau(\|x-p\|_{G}\,s - r\,s)$ & $-\tfrac12(x-\mu)^\top\Sigma^{-1}(x-\mu)$  \\
    Sharpness control & Separate temperature $\log\tau$ & Coupled to covariance (scale) \\
    Continuity model & Partition-of-unity weights & Kernel superposition \\
    Parameterization & det$(G)=1$ with explicit radius $r$ & Full covariance $\Sigma$ \\
    Initialization & Pixel-aligned or gradient-weighted & Gradient-weighted or saliency map \\
    Structure for compression & Induced adjacency graph & No inherent adjacency graph \\
    Acceleration & Top-$K$ propagation & {Tile binning} \\
    \bottomrule
  \end{tabular}
  }
  \vspace{0.5em}
  \caption{\textbf{Comparison of \name and Image-GS.} Key architectural and algorithmic differences between our Voronoi-based approach and Gaussian splatting methods.}
  \label{tab:comparison}
\end{table}

We match model sizes to target bitrates by adjusting grid resolutions, MLP width/depth, or the number of primitives as appropriate. All metrics are computed in linear color space.

We report bitrate as \emph{parameter-space} bits-per-pixel (BPP), i.e., the memory required to store the optimized representation divided by the number of image pixels. For \name, we compute BPP from the packed inference format used by our renderer: each site is encoded into 128 bits (uint4) with position stored as two 15-bit unorms over $[0,W{-}1]$ and $[0,H{-}1]$, color quantized to 11/11/10-bit unorms with per-image min/scale (colorR/G/B min + scale), $\log\tau$ and $r$ as 16-bit unorms with per-image min/scale, anisotropy direction as a 16-bit angle, and $\log a$ as fp16. The per-image quantization ranges (10 float scalars: logTauMin/Scale, radiusMin/Scale, colorR/G/B min/scale) are stored once and are negligible at typical resolutions. We thus compute
\[
\mathrm{BPP}=\frac{N_{\text{prim}}\cdot 16 \cdot 8}{H\cdot W}.
\]
For Image-GS~\cite{imagegs2025}, we assume 8 per-Gaussian parameters stored in fp16 (16 bytes): $\mu_x,\mu_y,\sigma_x,\sigma_y,\theta,c_r,c_g,c_b$ (no opacity term in their 2D model), and use the same accounting.

For Instant-NGP, which uses an implicit parameterization, we compute BPP from the total number of trainable parameters (hash-grid + MLP) and an assumed 16-bit storage precision (2 bytes per parameter), i.e.,
\[
\mathrm{BPP}=\frac{N_{\text{param}}\cdot 2 \cdot 8}{H\cdot W}.
\]

\begin{table}[t]
  \centering
  \small
    \resizebox{0.49\textwidth}{!}{
  \begin{tabular}{llcccc}
    \toprule
    \textbf{Method} & \textbf{Metric} & \textbf{0.2 BPP} & \textbf{0.3 BPP} & \textbf{0.4 BPP} & \textbf{0.5 BPP} \\
    \midrule
    Image-GS~\cite{imagegs2025} & PSNR$\uparrow$ & \cellcolor{secondcolor}31.32 & \cellcolor{secondcolor}32.79 & \cellcolor{secondcolor}33.80 & \cellcolor{secondcolor}34.57 \\
    & SSIM$\uparrow$ & \cellcolor{secondcolor}0.8923 & \cellcolor{secondcolor}0.9112 & \cellcolor{secondcolor}0.9228 & \cellcolor{secondcolor}0.9307 \\
    & LPIPS$\downarrow$ & \cellcolor{secondcolor}0.1309 & \cellcolor{secondcolor}0.1033 & \cellcolor{secondcolor}0.0873 & \cellcolor{secondcolor}0.0769 \\
    \midrule
    Instant-NGP~\cite{mueller2022instant} & PSNR$\uparrow$ & 26.66 & 29.41 & 29.86 & 30.69 \\
    & SSIM$\uparrow$ & 0.7703 & 0.8253 & 0.8304 & 0.8461 \\
    & LPIPS$\downarrow$ & 0.2472 & 0.1701 & 0.1656 & 0.1463 \\
    \midrule
    \textbf{\name (ours)} & PSNR$\uparrow$ & \cellcolor{bestcolor}\textbf{33.87} & \cellcolor{bestcolor}\textbf{35.72} & \cellcolor{bestcolor}\textbf{36.97} & \cellcolor{bestcolor}\textbf{37.86} \\
    & SSIM$\uparrow$ & \cellcolor{bestcolor}\textbf{0.8983} & \cellcolor{bestcolor}\textbf{0.9202} & \cellcolor{bestcolor}\textbf{0.9334} & \cellcolor{bestcolor}\textbf{0.9422} \\
    & LPIPS$\downarrow$ & \cellcolor{bestcolor}\textbf{0.0914} & \cellcolor{bestcolor}\textbf{0.0678} & \cellcolor{bestcolor}\textbf{0.0546} & \cellcolor{bestcolor}\textbf{0.0458} \\
    \bottomrule
  \end{tabular}
  }
  \vspace{0.5em}
  \caption{\textbf{Reconstruction quality on Image-GS dataset.} Average metrics over 45 images from the Image-GS benchmark~\cite{imagegs2025} at varying bitrates. \name outperforms baselines across all metrics and compression ratios.}
  \label{tab:image_gs_results}
\end{table}

\subsection{Image Compression Performance}

\begin{table}[t]
  \centering
  \small
    \resizebox{0.49\textwidth}{!}{
  \begin{tabular}{lccccc}
    \toprule
    \textbf{Method} & \textbf{PSNR$\uparrow$} & \textbf{SSIM$\uparrow$} & \textbf{LPIPS$\downarrow$} & \textbf{Time (s)$\downarrow$} \\
    \midrule
    G.Image (Chy.)~\cite{gaussianimage2024} & 39.36 & --- & --- & 13 \\
    G.Image (RS)~\cite{gaussianimage2024} & 39.78 & --- & --- & 14 \\
    Image-GS (g)~\cite{imagegs2025} & 39.04 & --- & --- & 28 \\
    Image-GS (S)~\cite{imagegs2025} & 38.65 & --- & --- & 28 \\
    Instant-GI~\cite{instantgi2025} & 41.41 & --- & --- & \cellcolor{secondcolor}10 \\

    {VBNF}~\cite{takikawa2022variable}  & {36.49} & {0.9609} & {0.0857} & {374} \\ 

    {DANF~\cite{belhe2023discontinuity}}  & {22.90} & {0.6176} & {0.3379} & {5.1} \\         
          {NFLD~\cite{liu20252d}} & {26.44} & {0.7032} & {0.3458} & {1700} \\ 
    
    Fast 2DGS~\cite{wang2025fast} & \cellcolor{secondcolor}43.13 & --- & --- & \cellcolor{secondcolor}10 \\
    \quad (w/o Positions) & 39.96 & --- & --- & \cellcolor{secondcolor}10 \\
    \quad (w/o Attributes) & 35.69 & --- & --- & \cellcolor{secondcolor}10 \\
    \quad (w/o Both) & 35.55 & --- & --- & \cellcolor{secondcolor}10 \\
    \midrule
    Image-GS~\cite{imagegs2025}$^\dagger$ & 36.90 & \cellcolor{secondcolor}0.9521 & 0.0272 & 28 \\
    Instant-NGP~\cite{mueller2022instant}$^\dagger$ & 37.72 & 0.9494 & \cellcolor{secondcolor}0.0249 & 8.2 \\
    \midrule
    \textbf{\name (ours)} & \cellcolor{bestcolor}\textbf{46.00} & \cellcolor{bestcolor}\textbf{0.9871} & \cellcolor{bestcolor}\textbf{0.0032} & \cellcolor{bestcolor}\textbf{2.2} \\
    \bottomrule
  \end{tabular}
  }
  \vspace{0.5em}
  \caption{\textbf{Image compression on Kodak.} Average reconstruction quality and training time over 24 images at $N=50{,}000$ primitives (approximately 16.0 BPP). Results for G.Image~\cite{gaussianimage2024}, Instant-GI~\cite{instantgi2025}, Fast 2DGS~\cite{wang2025fast}, and ablations are from Fast 2DGS paper (PSNR only). $^\dagger$We re-evaluated Image-GS and Instant-NGP under the same protocol with all metrics. For Instant-GI, $N=47{,}246$.}
  \label{tab:kodak_compression}
\end{table}

On Kodak dataset~\cite{kodak} (Table~\ref{tab:kodak_compression}), at $N=50{,}000$ primitives ($\approx 16.0$ BPP), \name reaches 46.00 dB PSNR (0.9871 SSIM, 0.0032 LPIPS) with a 2.2s encoding time. Under the same protocol, it improves over Fast 2DGS by +2.87 dB (43.13 dB, PSNR-only) while being $4.5\times$ faster (10s $\rightarrow$ 2.2s), and also outperforms the re-evaluated Image-GS and Instant-NGP baselines in both quality and encoding cost. {For completeness, Table~\ref{tab:kodak_compression} also reports DANF and NFLD as discontinuity-aware references. Their intended use cases differ from that of \name: DANF assumes known discontinuity curves, whereas NFLD learns discontinuities on a triangle mesh initialized from detected edges. In contrast, \name captures sharp, content-aligned discontinuities without prescribed curves or edge-to-mesh preprocessing, while retaining adaptive budget control and fixed-size per-pixel top-$K$ evaluation for efficient fitting. We also compare against Variable Bitrate Neural Fields (VBNF)~\cite{takikawa2022variable}, a variable-bitrate neural-field baseline based on vector-quantized feature grids. SAD outperforms VBNF in both quality and encoding time, including on Kodak and at both evaluated bitrates on DIV2K (Table~\ref{tab:div2k_results}).}

Across datasets and bitrates (Tables~\ref{tab:image_gs_results}--\ref{tab:clic_results}), \name shows consistent rate--distortion gains: +2.55--3.29 dB over Image-GS on the Image-GS benchmark (0.2--0.5 BPP), +1.52/+2.58 dB on DIV2K, and +1.17/+1.98 dB on CLIC at 0.5/2.0 BPP. {These gains stem from the structure of the representation: the explicit soft partition provides localized ownership and content-aligned transitions, while the anisotropic metric with separate temperature control enables sharper discontinuities and tighter spatial adaptation than isotropic Gaussian kernels under a fixed-size local query budget.}

{In this paper, we focus on \emph{parametric} compression in a narrower sense: the image is represented by optimized site parameters that remain directly usable inside a differentiable renderer. Accordingly, the BPP reported here measures parameter-space storage of the representation and should be read as a proxy for representational compactness within this setting, not as a claim that SAD is a replacement for mature production codecs such as JPEG2000 or WebP. Those codecs target a different operating point, combining transform design, entropy coding, and deployment-oriented decoding pipelines. Our empirical comparisons are therefore against differentiable image representations with similar per-instance optimization and random-access goals. Entropy coding of SAD parameters is an important future direction, but not a completely orthogonal one: the achievable gains will depend on the statistics of the learned sites and their induced adjacency. We nevertheless expect the explicit site layout to provide useful handles for prediction and coding, e.g., neighbor-conditioned differential coding or graph-aware quantization.}

\begin{table}[t]
  \centering
  \small
      \resizebox{0.49\textwidth}{!}{
  \begin{tabular}{llcc}
    \toprule
    \textbf{Method} & \textbf{Metric} & \textbf{0.5 BPP} & \textbf{2.0 BPP} \\
    \midrule
    Image-GS~\cite{imagegs2025} & PSNR$\uparrow$ & \cellcolor{secondcolor}28.48 & 32.15 \\
    & SSIM$\uparrow$ & \cellcolor{secondcolor}0.7914 & \cellcolor{secondcolor}0.8820 \\
    & LPIPS$\downarrow$ & \cellcolor{secondcolor}0.2515 & \cellcolor{secondcolor}0.1480 \\
    \midrule
    Instant-NGP~\cite{mueller2022instant} & PSNR$\uparrow$ & 26.44 & 29.24 \\
    & SSIM$\uparrow$ & 0.7045 & 0.7940 \\
    & LPIPS$\downarrow$ & 0.2778 & 0.1755 \\
    \midrule

      {VBNF}~\cite{takikawa2022variable} & PSNR$\uparrow$ & {27.13} & {31.28}  \\
    & SSIM$\uparrow$ & {0.7495} & {0.8737} \\
    & LPIPS$\downarrow$ & {0.4321} & {0.2765} \\
    \midrule

    Instant-GI~\cite{instantgi2025}$^*$ & PSNR$\uparrow$ & --- & \cellcolor{bestcolor}38.01 \\
    Fast 2DGS~\cite{wang2025fast}$^*$ & PSNR$\uparrow$ & --- & \cellcolor{secondcolor}37.81 \\
    \midrule
    \textbf{\name (ours)} & PSNR$\uparrow$ & \cellcolor{bestcolor}\textbf{30.00} & 34.73 \\
    & SSIM$\uparrow$ & \cellcolor{bestcolor}\textbf{0.7982} & \cellcolor{bestcolor}\textbf{0.9115} \\
    & LPIPS$\downarrow$ & \cellcolor{bestcolor}\textbf{0.1995} & \cellcolor{bestcolor}\textbf{0.0844} \\
    \bottomrule
  \end{tabular}
  }
  \vspace{0.5em}
  \caption{\textbf{Reconstruction quality on DIV2K validation set.} Average metrics over 100 images at varying bitrates and variable resolution (BPP computed per-image from its H×W). $^*$Results from Fast 2DGS~\cite{wang2025fast} use $N=50{,}000$ primitives (approximately 2.0 BPP) and require pre-trained priors (PSNR only).}
  \label{tab:div2k_results}
\end{table}

\begin{figure*}[t]
  \centering
  \includegraphics[width=\textwidth]{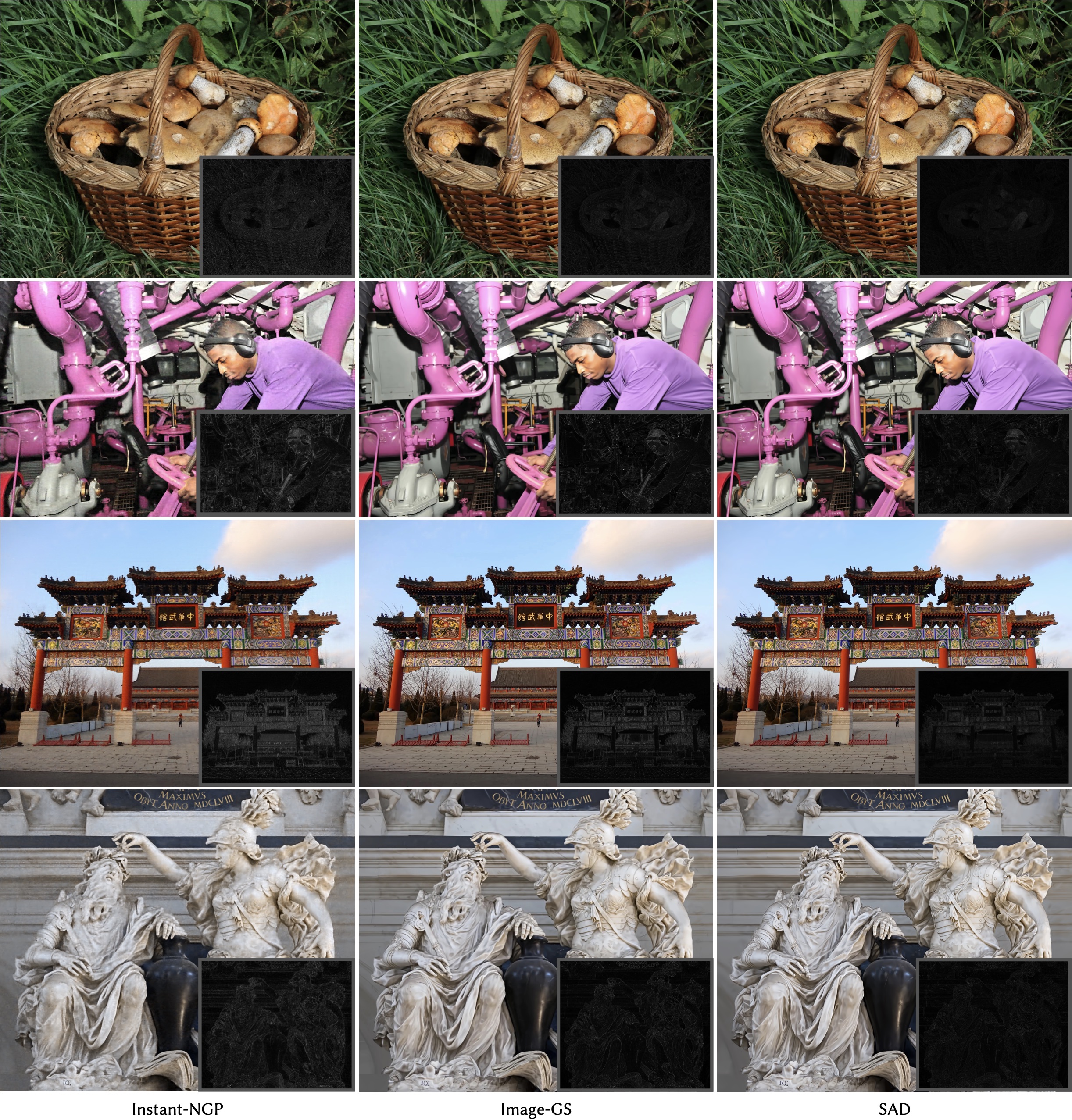}
  \vspace{-7mm}
  \caption{\textbf{Qualitative comparison on DIV2K.} Results for Instant-NGP, Image-GS, and \name; error maps are shown at the bottom-right of each result.}
  \label{fig:div2k_qual}
  \vspace{-4mm}
\end{figure*}

\begin{table}[t]
  \centering
  \small
      \resizebox{0.49\textwidth}{!}{
  \begin{tabular}{llcc}
    \toprule
    \textbf{Method} & \textbf{Metric} & \textbf{0.5 BPP} & \textbf{2.0 BPP} \\
    \midrule
    Image-GS~\cite{imagegs2025} & PSNR$\uparrow$ & \cellcolor{secondcolor}30.65 & \cellcolor{secondcolor}34.15 \\
    & SSIM$\uparrow$ & \cellcolor{bestcolor}0.8223 & \cellcolor{secondcolor}0.8907 \\
    & LPIPS$\downarrow$ & \cellcolor{secondcolor}0.2280 & 0.1449 \\
    \midrule
    Instant-NGP~\cite{mueller2022instant} & PSNR$\uparrow$ & 28.59 & 32.67 \\
    & SSIM$\uparrow$ & 0.7559 & 0.8475 \\
    & LPIPS$\downarrow$ & 0.2351 & \cellcolor{secondcolor}0.1287 \\
    \midrule
    \textbf{\name (ours)} & PSNR$\uparrow$ & \cellcolor{bestcolor}\textbf{31.82} & \cellcolor{bestcolor}\textbf{36.13} \\
    & SSIM$\uparrow$ & \cellcolor{secondcolor}\textbf{0.8176} & \cellcolor{bestcolor}\textbf{0.9112} \\
    & LPIPS$\downarrow$ & \cellcolor{bestcolor}\textbf{0.1870} & \cellcolor{bestcolor}\textbf{0.0884} \\
    \bottomrule
  \end{tabular}
  }
  \vspace{0.5em}
  \caption{\textbf{Reconstruction quality on CLIC validation set.} Average metrics over 41 images at varying bitrates.}
  \label{tab:clic_results}
\end{table}

\subsection{Training Performance}

To isolate per-iteration computational cost from convergence behavior, we benchmark training performance on a controlled test set with consistent epoch definitions across all methods. We define one \emph{epoch} as a single pass over all pixels in the image.

\paragraph{Experimental protocol.}
We created five test images at resolutions 512$^2$, 768$^2$, 1024$^2$, 1536$^2$, and 2048$^2$ by downscaling a single source image from the Image-GS dataset. Each method was trained at 1.0 BPP with three runs per resolution on an NVIDIA RTX 5090 GPU. For \name, we used default settings (4000 iterations) with CUDA backend. For Image-GS~\cite{imagegs2025}, we used 5000 training steps with the number of Gaussians set to achieve 1.0 BPP.

For Instant-NGP~\cite{mueller2022instant}, standard practice uses a fixed batch size (262{,}144 random UV samples per step), which results in dramatically different sample coverage across resolutions: at 512$^2$ each pixel is sampled once per step on average, while at 2048$^2$ each pixel is sampled only once every 16 steps. To enable consistent per-epoch comparison, \textit{we modified Instant-NGP to set \texttt{training\_batch\_size} equal to the total number of pixels ($\text{width} \times \text{height}$)}, ensuring each training step constitutes exactly one full epoch over the image. We trained for 10{,}000 steps with this configuration. Under these settings, all three methods process the same total number of pixel samples per epoch.

\begin{figure}[t]
  \centering
  \includegraphics[width=1.0\linewidth]{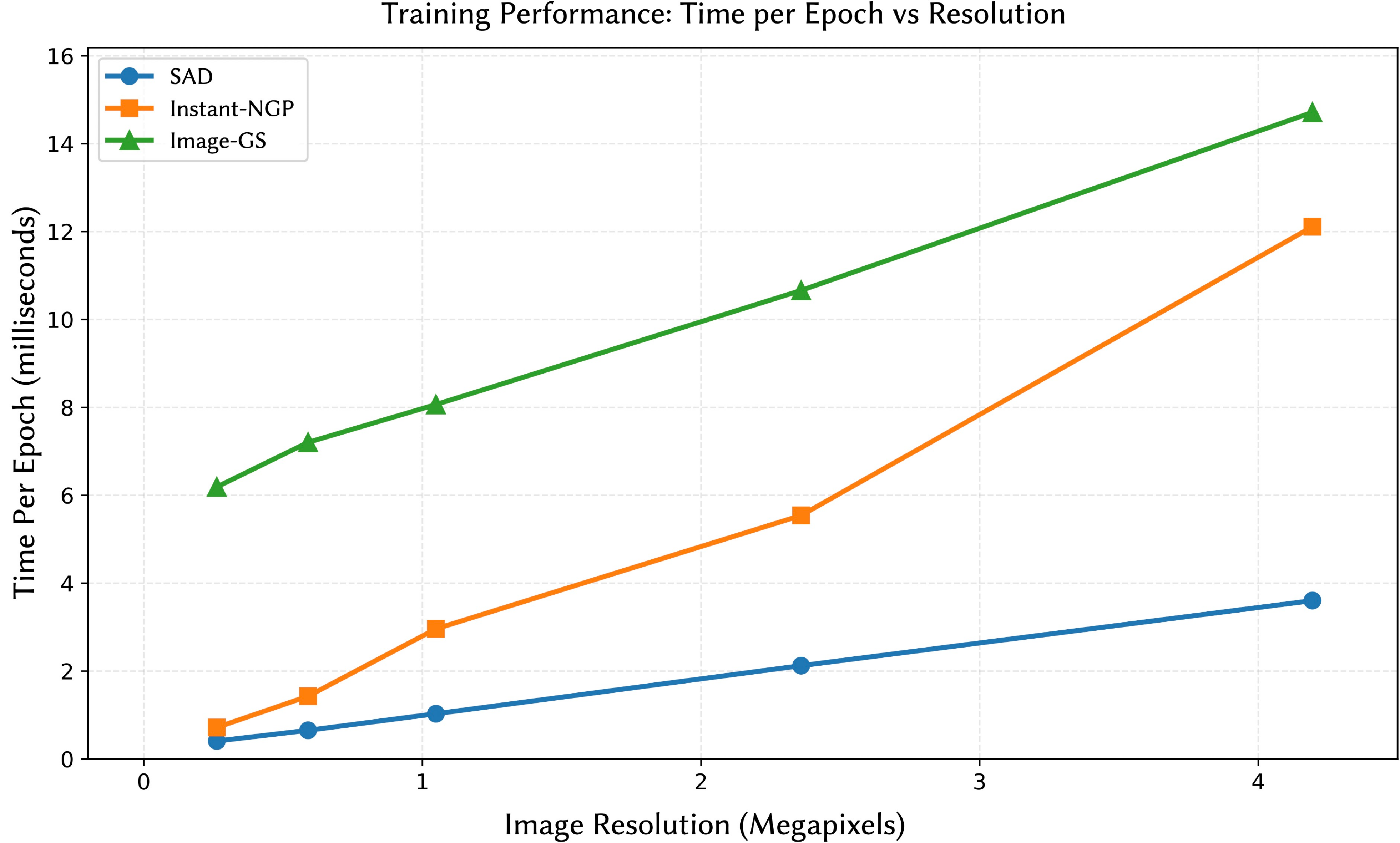}
  \caption{\textbf{Training performance: time per epoch vs image resolution.} We report time per epoch (one full pass over all pixels) across methods. All methods were trained at 1.0 BPP on five test resolutions (512$^2$ to 2048$^2$) with three runs each. For Instant-NGP, we set batch size equal to total pixels to ensure each step processes exactly one epoch; standard NGP uses a fixed 262k batch size, resulting in vastly different sample coverage across resolutions. \name is 1.75--3.36$\times$ faster per epoch than Instant-NGP and 4.08--15.10$\times$ faster than Image-GS. Results averaged over three runs with negligible variance (std dev $<$2\% of mean). NVIDIA RTX 5090.}
  \label{fig:training_time_vs_resolution}
\end{figure}

\paragraph{Results.}
Figure~\ref{fig:training_time_vs_resolution} shows time per epoch as a function of image resolution. \name is the fastest method per epoch across all resolutions, achieving 1.75--3.36$\times$ speedup over Instant-NGP and 4.08--15.10$\times$ speedup over Image-GS. At 2k resolution (4.19 MP), \name processes one epoch in 3.6 ms compared to 12.1 ms for Instant-NGP and 14.7 ms for Image-GS. All methods exhibit near-linear scaling with resolution, as expected when processing all pixels per epoch.

Critically, \name not only runs faster per epoch but also attains higher final quality under the official step budgets (Image-GS 5{,}000, Instant-NGP 10{,}000, ours 4{,}000). As shown in our convergence analysis (Section~\ref{sec:convergence}), \name achieves strong reconstruction quality (37.5 dB PSNR on 2k images at 0.5 BPP) in 2000 iterations. Combining faster per-epoch execution with these step budgets, under the equal-sample regime (full-image batch for Instant-NGP), \name achieves 4--8$\times$ end-to-end speedup over Instant-NGP and 5--19$\times$ speedup over Image-GS in total wall-clock training time. {
This wall-clock advantage is not due only to a lower iteration count, but also to representation-enabled implementation choices: reusing the previous top-$K$ lists avoids full per-pixel candidate search, and our tiled threadgroup-hash reduction substantially reduces the cost of gradient accumulation, making the overall speedup a representation--implementation co-design effect.}

\subsection{Convergence Analysis}
\label{sec:convergence}

Unlike Gaussian splatting approaches that start with few primitives and progressively add more through densification, \name begins with a large initial site count and progressively removes sites with low contribution through adaptive pruning. To study the convergence behavior and training efficiency, we trained \name on three resolution/bitrate configurations with varying iteration counts (500, 1000, 2000, 3000, 4000, 5000): (1) five diverse 2k images (2048$\times$2048) from the Image-GS dataset~\cite{imagegs2025} at 0.5 BPP, (2) the same five images downscaled to 1k (1024$\times$1024) at 1.0 BPP, and (3) five Kodak images~\cite{kodak} (768$\times$512) at 4.0 BPP. Importantly, the final site count remains consistent across all iteration variants within each configuration because our target BPP calculation automatically adjusts the prune schedule based on the total iteration budget—longer training uses more aggressive early pruning to reach the same final primitive count.

\begin{figure}[t]
  \centering
  \includegraphics[width=1.0\linewidth]{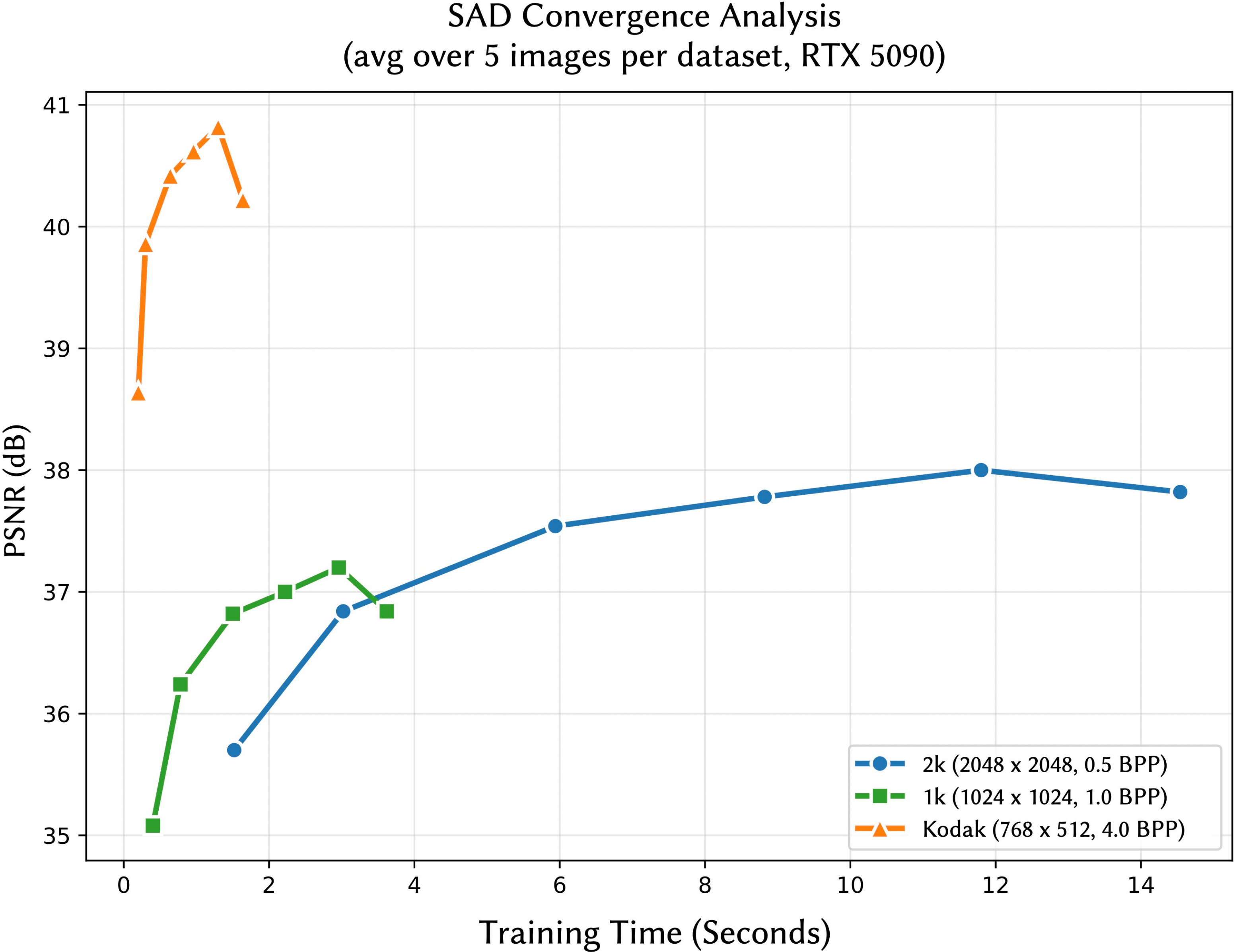}
  \caption{\textbf{Convergence analysis.} Average PSNR vs training time across three resolution/bitrate configurations. All show similar convergence patterns: quality plateaus around half the maximum training time (2000 iterations), with minimal improvement or slight degradation from longer training.}
  \label{fig:convergence}
\end{figure}

{Figure~\ref{fig:convergence} shows the average PSNR as a function of training time for all three configurations. All curves exhibit rapid initial improvement followed by diminishing returns, with a consistent pattern across resolutions and bitrates. For 2k images (0.5 BPP), training for 2000 iterations ($\sim$6s) achieves 37.5 dB, nearly matching the peak of 38.0 dB at 4000 iterations ($\sim$12s), with degradation to 37.8 dB at 5000 iterations. For 1k images (1.0 BPP), 2000 iterations ($\sim$1.5s) achieves 36.8 dB, approaching the peak of 37.2 dB at 4000 iterations ($\sim$3s), with degradation to 36.8 dB at 5000 iterations. For Kodak images (4.0 BPP), 2000 iterations ($\sim$0.6s) achieves 40.4 dB, approaching the peak of 40.8 dB at 4000 iterations ($\sim$1.3s), with degradation to 40.2 dB at 5000 iterations. This behavior is a schedule-induced optimization effect under a fixed-BPP constraint, not classical overfitting: changing the total iteration budget changes the densify/prune trajectory required to reach the same final site count, which can lead to a slightly worse local optimum after saturation. Critically, \name achieves strong reconstruction quality in approximately half the time of full training across all configurations, making it well-suited for high-throughput scenarios.}

\begin{figure*}
  \centering
  \includegraphics[width=\textwidth]{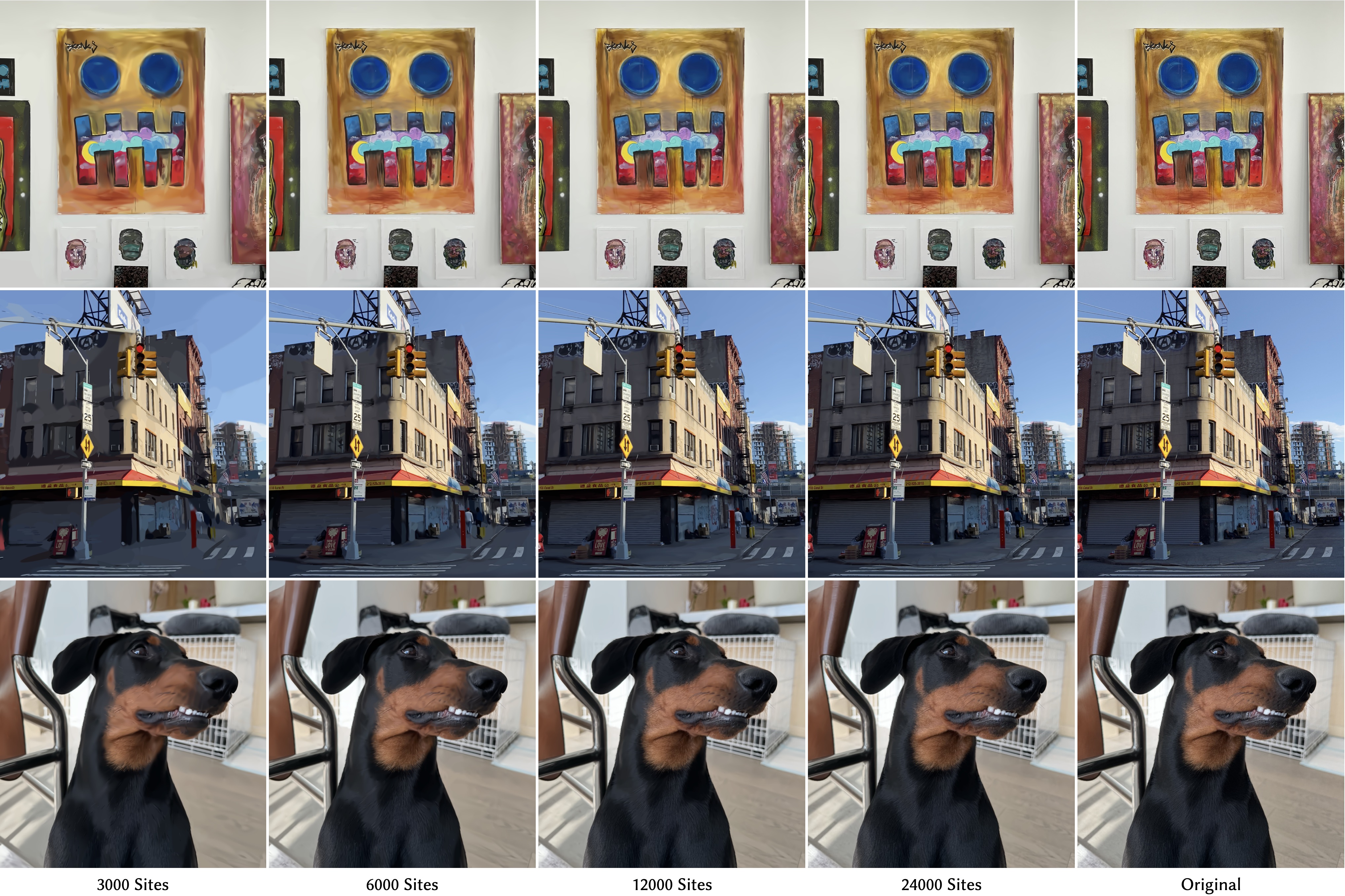}
  \caption{\textbf{Effect of site budget on reconstruction quality.} Three 2048$\times$1900 DIY photos rendered by \name at 3k/6k/12k/24k sites (left to right), with the original in the final column.}
  \label{fig:sites_count}
\end{figure*}

\subsection{{Top-$K$ Propagation Convergence}}
We evaluate the convergence of the top-$K$ propagation by comparing the approximate top-$K$ lists against exact top-$K$ on 256 random pixels in Table~\ref{tab:vpt_convergence}. Site centers are initialized without collisions by sampling unique pixel cells and adding layered subpixel offsets. We report \emph{Top-8 exact match}: the fraction
of pixels whose top-$8$ set exactly matches the ground-truth top-$8$ under the
rendering score, for the default propagation settings, averaged over 4 trials.

\begin{table}[t]
  \centering
  \small
  \begin{tabular}{lccc}
    \toprule
    Domain, Sites & 8 passes & 12 passes & 16 passes \\
    \midrule
    \multicolumn{4}{l}{\textit{1024$^2$}} \\
    16k & 0.963 & 0.970 & 0.970 \\
    65k & 0.398 & 0.867 & 0.867 \\
    131k & 0.072 & 0.782 & 0.782 \\
    \midrule
    \multicolumn{4}{l}{{\textit{2048$^2$}}} \\
    16k & 0.981 & 0.987 & 0.987 \\
    65k & 0.393 & 0.956 & 0.956 \\
    131k & 0.059 & 0.907 & 0.907 \\
    \bottomrule
  \end{tabular}
  \vspace{0.5em}
  \caption{\textbf{Top-$K$ propagation convergence (Top-8 exact match).} Sites are in thousands (e.g., 16k = 16{,}384). Values are averages over 4 trials with 256 random pixels per image.}
  \label{tab:vpt_convergence}
\end{table}

In this experiment, across 1k--2k domains, Top-8 exact match rises steeply through 8 passes and largely
saturates by 12--16. We use a single pass per training step; for rendering, we treat 8 passes as coherent and report 12--16 to show convergence.

\subsection{Rendering Performance}
\paragraph{Protocol.}
We measure GPU time per render call on an NVIDIA RTX 5090 at 512$^2$, 1024$^2$,
and 2048$^2$. We keep the 1.0 BPP setting used in the training benchmark
(primitive counts set by the same parameter-space BPP accounting as above). For \name, we separate
top-$K$ propagation updates from rendering. Reported \name render times include
1 JFA round + 16 propagation passes + render, which is the setting used for
near-converged rendering; training uses a single propagation pass per step.
Image-GS and Instant-NGP have no reusable top-$K$ propagation, so their per-call cost is the full render. All timings use 10 warmup and 100 timed runs, measured
with GPU events and excluding CPU readback or image saving.

\begin{table}[t]
  \centering
  \small
  \begin{tabular}{lccc}
    \toprule
    {Method} & {512$^2$} & {1024$^2$} & {2048$^2$} \\
    \midrule
    {Instant-NGP} & {0.405} & {0.411} & {0.444} \\
    {Image-GS} & {0.625} & {0.953} & {2.571} \\
    {\textbf{\name}} & {1.100} & {4.614} & {26.539} \\
    \bottomrule
  \end{tabular}
  \vspace{0.5em}
  \caption{\textbf{Render time per call (ms).} Full GPU render time for each method. For \name, this includes JFA seeding, 16 propagation passes, and rendering.}
  \label{tab:render_call}
\end{table}

{Table~\ref{tab:render_breakdown} decomposes \name into its update and render
stages. The gap between the full refresh (Table~\ref{tab:render_call}) and the
update-only cost corresponds to the one-time JFA seed. A single-pass update
approximates the incremental update used during training iterations.}

\begin{table}[t]
  \centering
  \small
  \begin{tabular}{lccc}
    \toprule
    {Stage (\name)} & {512$^2$} & {1024$^2$} & {2048$^2$} \\
    \midrule
    {Render only (cached top-$K$ list)} & {0.015} & {0.034} & {0.132} \\
    {Update only (16 passes)} & {0.728} & {2.568} & {16.009} \\
    {Update only (1 pass, training)} & {0.053} & {0.168} & {1.006} \\
    \bottomrule
  \end{tabular}
  \vspace{0.5em}
  \caption{{\textbf{\name rendering breakdown (ms).} Top-$K$ propagation update and render costs are reported separately.}}
  \label{tab:render_breakdown}
\end{table}

{These measurements expose three regimes for \name: render-only (cached
top-$K$ list), incremental update (1 pass, typical during training), and full
refresh (16 passes after large edits or re-initialization). In practice, a
single-pass update produces a coherent preview almost immediately, while
additional passes progressively refine boundaries and fine detail. For pure
image-space zoom/pan, the rendered image can be resampled without re-rendering; generating a new pixel grid (true re-render at a different resolution) requires
rebuilding the top-$K$ list. Fast random access corresponds to the render-only
path with cached top-$K$ lists; a refresh incurs the update costs above. During training, candidate updates are amortized through single-pass temporal updates, and the total cost is dominated by the combined forward, backward, and gradient-accumulation kernels rather than full candidate refreshes.}

\subsection{Ablation Study on Site Count }
In Figure~\ref{fig:sites_count}, we visualize reconstruction quality for three DIY photos at increasing site budgets (3k, 6k, 12k, 24k), followed by the original. Inputs are 2048$\times$1900 and are not part of a benchmark dataset. The global layout and dominant colors are stable even at 3k sites, while additional budget primarily restores mid- and high-frequency detail (thin structures, text, brush strokes, wire grids, and fur strands). Visual gains are most apparent from 3k to 12k, with diminishing returns by 24k as remaining improvements are fine texture and edge crispness.
All examples use the default training configuration with a target BPP chosen to reach the shown site counts. Training starts from 128k initialized sites and uses the adaptive budget mechanism from \S\ref{sec:method}: densify splits the highest error-density sites ($s_i$), while prune removes the lowest removal-delta sites; the prune/densify percentiles are automatically rescaled to meet the target count under the fixed iteration budget.

\subsection{{Ablation Study on Learnable Parameters}}
\label{sec:ablation}
We conduct a comprehensive ablation study to evaluate the contribution of each learnable parameter in our representation. Starting from a baseline with fixed temperature ($\log\tau$), we progressively enable optimization of additional parameters: adaptive temperature, additive radius, and anisotropy (direction and magnitude). All experiments use identical training settings on 5 test images at 2048$\times$2048 resolution at 0.5 BPP.

\begin{figure}[t]
  \centering
  \includegraphics[width=\linewidth]{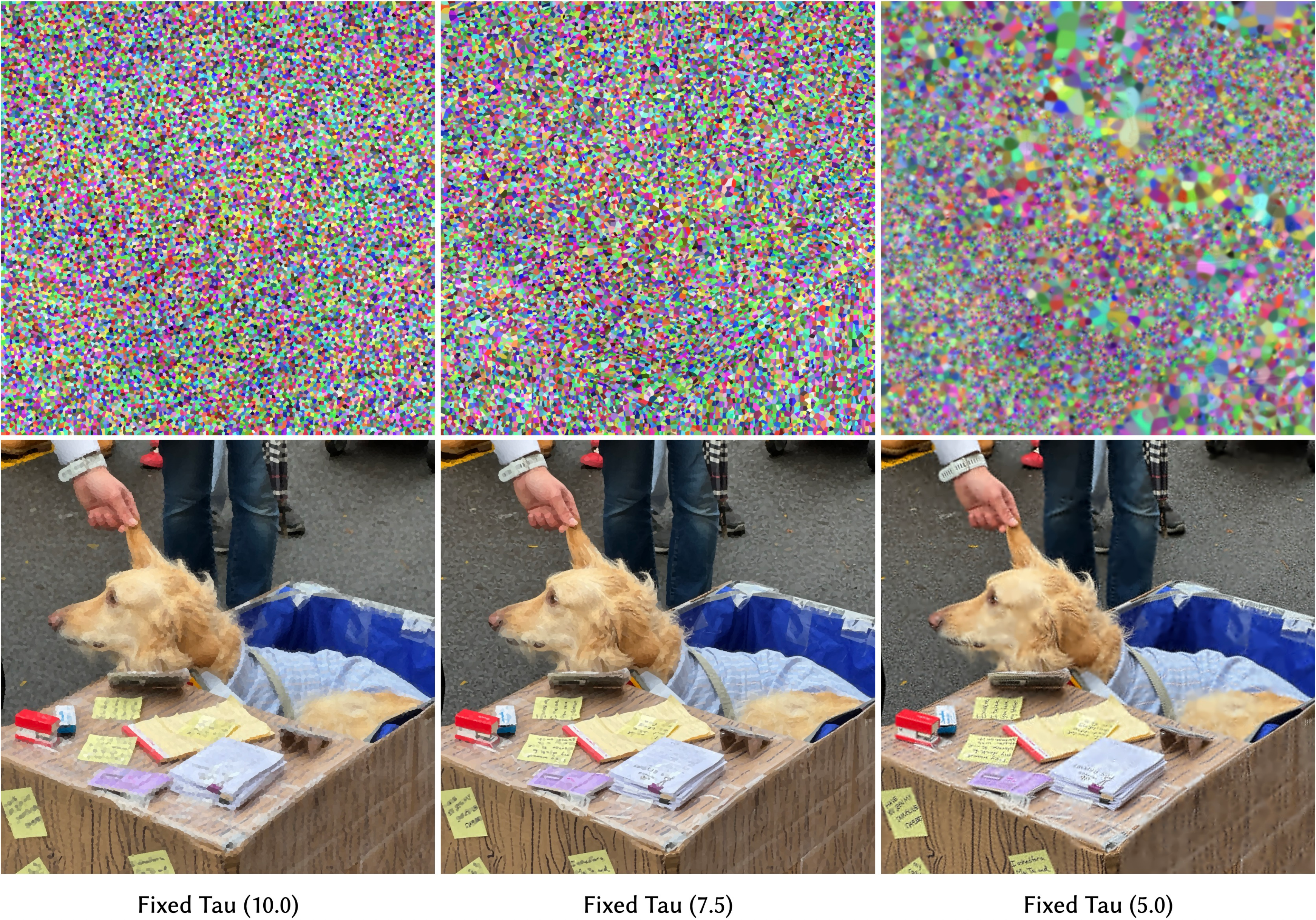}
  \caption{\textbf{Effect of fixed temperature on reconstruction.} From left to right: $\log\tau=10.0$ (23.96 dB), $\log\tau=7.5$ (25.12 dB), $\log\tau=5.0$ (25.37 dB). Higher temperatures produce sharper boundaries while lower temperatures create softer blending.}
  \label{fig:fix_tau_comparison}
\end{figure}

\paragraph{Fixed Temperature Baseline.}
Figure~\ref{fig:fix_tau_comparison} shows reconstruction quality across different fixed temperature values ($\log\tau \in \{5.0, 7.5, 10.0\}$). Higher temperatures produce sharper cell boundaries suited for edges, while lower temperatures enable softer blending for smooth gradients. No single fixed temperature achieves optimal results across all image regions, motivating adaptive temperature learning.
\begin{figure*}[t]
  \centering
  \includegraphics[width=\linewidth]{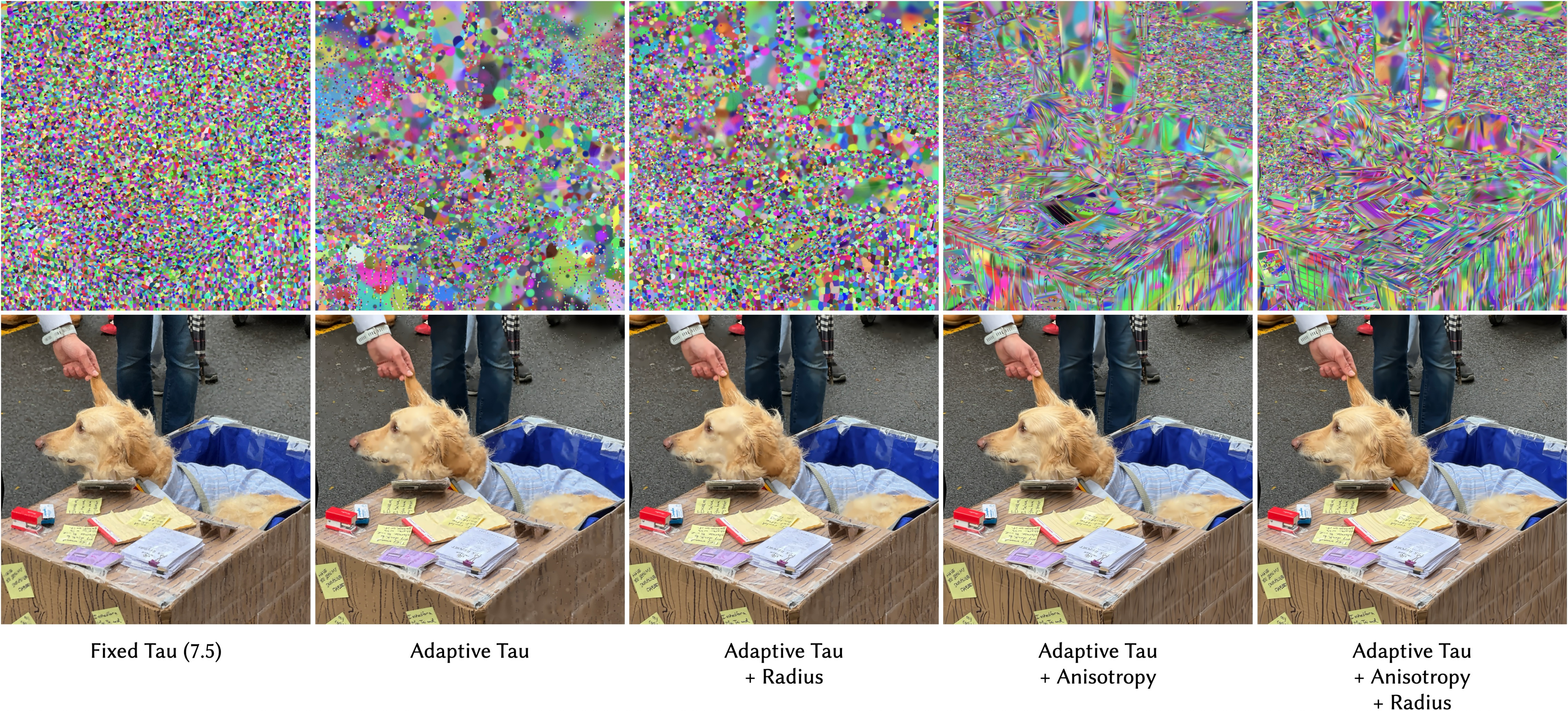}
  \vspace{-7mm}
  \caption{\textbf{Effect of learnable parameters on cell structure and reconstruction.} Top row: \name diagrams. Bottom row: Reconstructed images. From left to right: Fixed $\log\tau=7.5$, Adaptive Tau, +Radius, +Anisotropy, Full model (+Radius+Anisotropy). Anisotropic cells create characteristic spiral patterns in smooth regions and align with edges in detailed areas.}
  \label{fig:parameter_analysis}
  \vspace{-3mm}
\end{figure*}

\paragraph{Parameter Ablation.}
Table~\ref{tab:ablation} quantifies the contribution of each learnable parameter. Enabling adaptive temperature improves PSNR by +2.30 dB over the best fixed baseline (28.20 dB $\rightarrow$ 30.50 dB), allowing sites to locally adjust their influence regions. Adding a learnable radius provides +1.26 dB additional improvement (30.50 dB $\rightarrow$ 31.76 dB) by enabling sites to expand or contract their effective coverage. The most substantial gain comes from anisotropy: adding directional adaptation to adaptive temperature yields +4.27 dB (30.50 dB $\rightarrow$ 34.77 dB), demonstrating that elongated cells aligned with image gradients dramatically improve reconstruction of edges and directional textures. The full model combining all parameters achieves 35.35 dB, a +7.15 dB improvement over the fixed baseline.

\begin{table}[t]
  \centering
  \small
  \begin{tabular}{lcccc}
    \toprule
    \textbf{Configuration} & \textbf{Tau} & \textbf{Radius} & \textbf{Aniso} & \textbf{PSNR$\uparrow$} \\
    \midrule
    Fixed $\log\tau=5.0$ & \xmark & \xmark & \xmark & 27.84 \\
    Fixed $\log\tau=7.5$ & \xmark & \xmark & \xmark & 28.20 \\
    Fixed $\log\tau=10.0$ & \xmark & \xmark & \xmark & 26.02 \\
    \midrule
    Adaptive Tau & \cmark & \xmark & \xmark & 30.50 \\
    + Radius & \cmark & \cmark & \xmark & 31.76 \\
    + Anisotropy & \cmark & \xmark & \cmark & 34.77 \\
    \textbf{Full \name} & \cmark & \cmark & \cmark & \cellcolor{bestcolor}\textbf{35.35} \\
    \bottomrule
  \end{tabular}
  \vspace{0.5em}
  \caption{\textbf{Ablation study: contribution of learnable parameters.} Average PSNR over 5 test images (2048$\times$2048). Checkmarks indicate parameters with non-zero learning rate. Adaptive temperature provides +2.30 dB over fixed baseline; anisotropy contributes the largest single improvement (+4.27 dB over adaptive tau alone).}
  \label{tab:ablation}
\end{table}

\paragraph{Visual Analysis.}
Figure~\ref{fig:parameter_analysis} visualizes the effect of each parameter combination. The diagrams (top row) reveal qualitatively different cell structures: fixed temperature produces uniform cell sizes, adaptive temperature allows local size variation, adding radius enables expansion of important cells, and anisotropy creates elongated cells that align with image edges. These structural differences directly translate to reconstruction quality, with the full model producing the sharpest edges and most accurate textures.

\section{More Applications}

\subsection{Differentiable PDE Solving}
\begin{figure}[t]
  \centering
  \includegraphics[width=\linewidth]{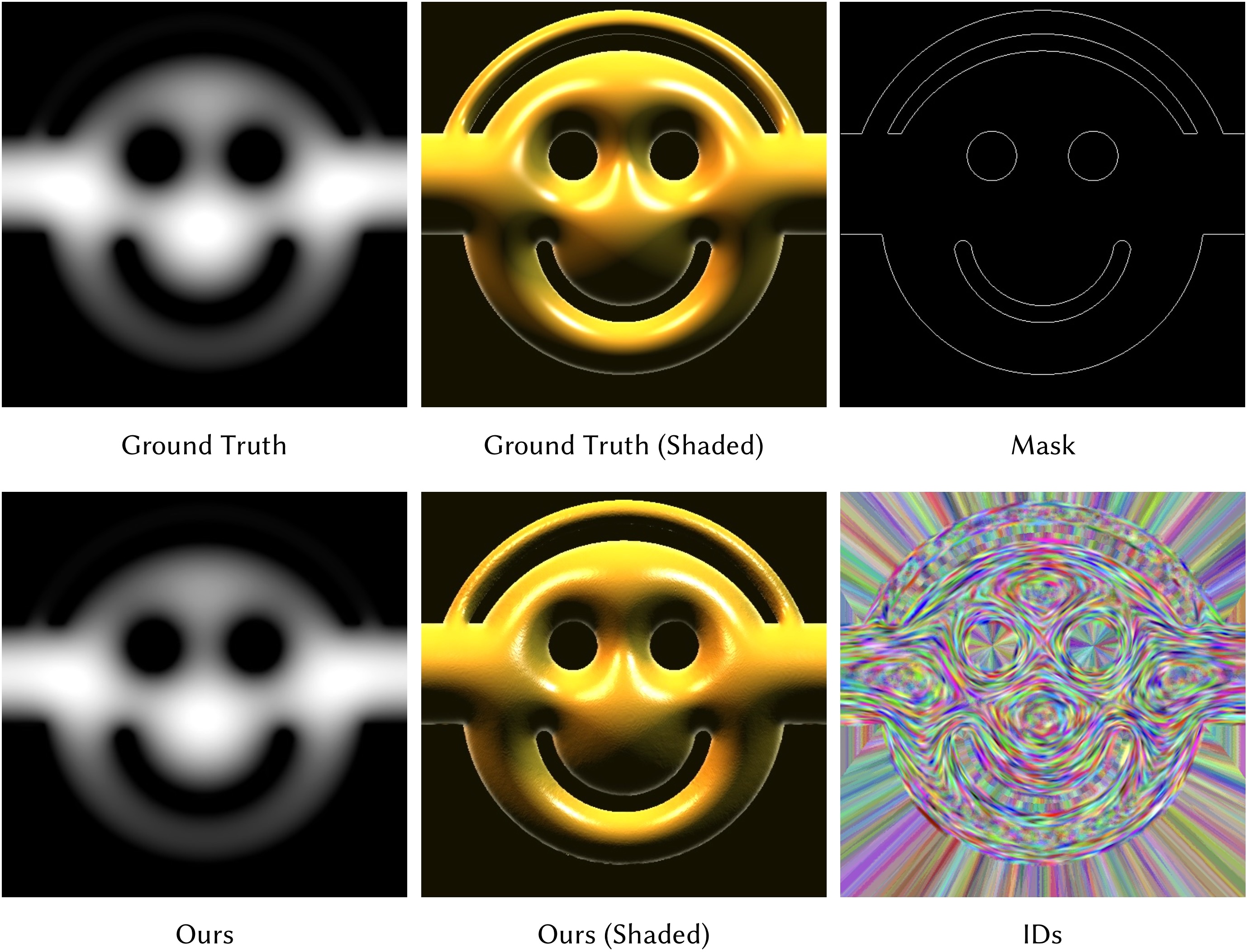}
  \caption{\textbf{Differentiable Poisson solving on irregular domain.} Top: Ground truth solution, shaded heightfield visualization, and binary mask defining the smiley domain. Bottom: Our reconstructed solution, shaded output, and learned site IDs. The colorful ID map reveals adaptive site concentration near boundaries and high-curvature features. Explicit sites enable hard constraint enforcement by freezing boundary sites during optimization.}
  \label{fig:poisson}
\end{figure}

Beyond image fitting, \name serves as an adaptive mesh-free representation for differentiable physics simulation. We demonstrate this by solving the 2D Poisson equation $\nabla^2 u = f$ on an irregular domain $\Omega$ (smiley mask, 512$\times$512), with Dirichlet boundary conditions $u = 0$ on $\partial\Omega$ and a spatially constant source term $f(\mathbf{x}) = -4$ for $\mathbf{x} \in \Omega$. We initialize 20k interior sites (optimized via gradient descent on the PDE residual) and 3k boundary sites placed along the zero-level contour. Critically, the explicit site structure allows us to \emph{freeze} boundary sites after initialization, enforcing hard Dirichlet constraints by simply excluding them from gradient updates—a straightforward operation that is non-trivial in implicit MLP-based representations~\cite{sitzmann2020implicit} where satisfying hard boundary conditions typically requires specialized distance-function constructions~\cite{kraus2024sdfpinn} or penalty-based soft constraints. With this setup, our representation converges to machine precision (MSE $< 10^{-6}$) in 1000--2000 gradient steps. The learned site distribution (Figure~\ref{fig:poisson}) concentrates naturally near boundaries and high-curvature regions, demonstrating content-adaptive behavior. This shows that \name can serve as a general differentiable spatial representation for physics-informed optimization, with explicit structure enabling direct constraint enforcement and localized control.

\subsection{1D Signal Fitting}
\begin{figure}[t]
  \centering
  \includegraphics[width=\linewidth]{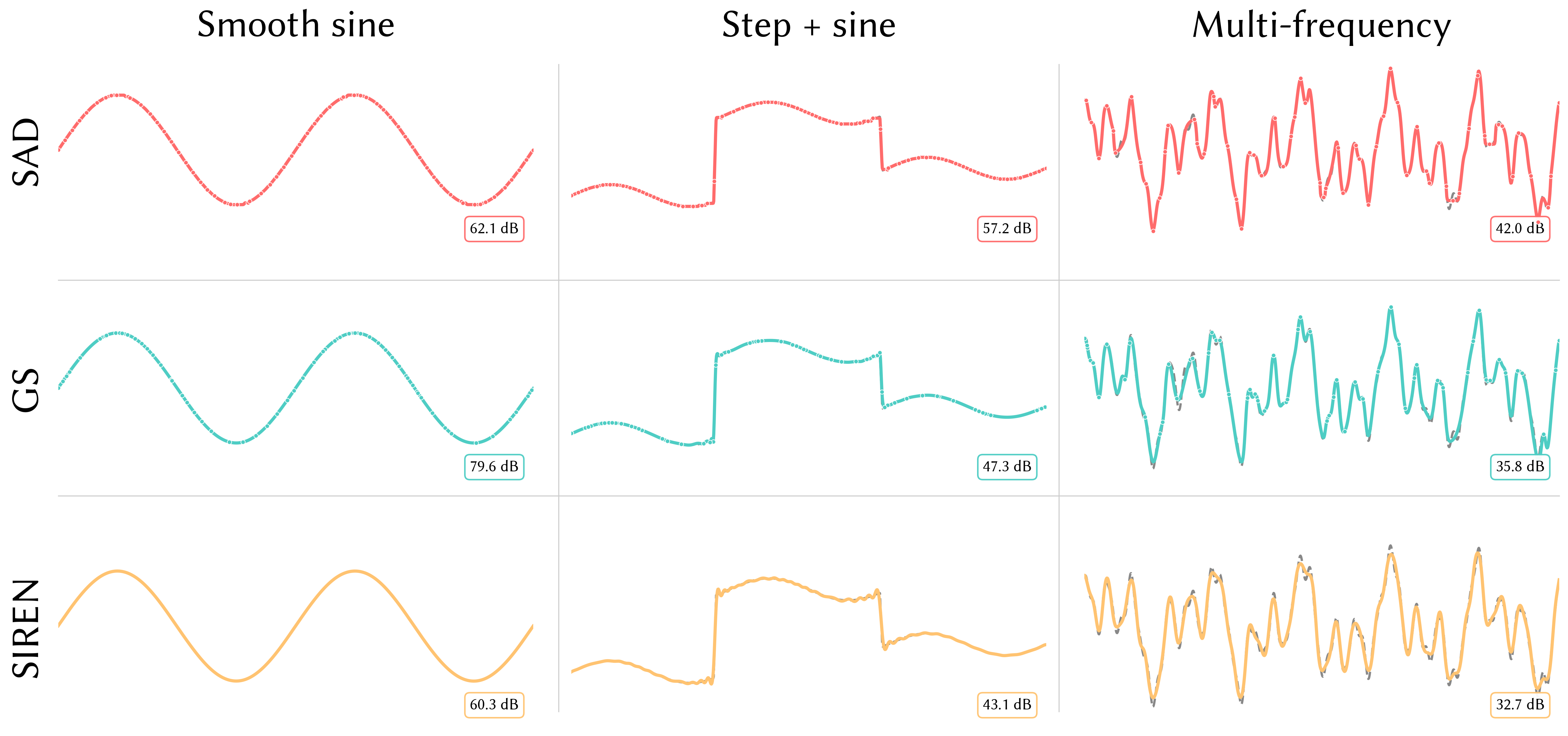}
  \caption{{\textbf{1D signal fitting at a matched 256-parameter budget.}
  Rows: \name~(64 sites $\times$ 4 params), Gaussian splatting~\cite{kerbl20233d}
  (64 splats $\times$ 4 params), and SIREN~\cite{sitzmann2020implicit}
  ($[1,16,16,1]$, $\omega_0{=}15$, 321 params).
  Columns: smooth sine, step-plus-sine, and multi-frequency periodic target.
  Dots on the \name and Gaussian rows mark primitive positions.
  Gaussian splatting fits smooth regions near-optimally but rounds off the step;
  SIREN encodes frequency content globally yet exhibits Gibbs-like ringing at the discontinuity;
  \name's learnable per-site temperature sharpens the partition exactly at the step while
  remaining smooth elsewhere.}}
  \label{fig:1d_signal}
\end{figure}
{
\name extends naturally across low-dimensional signal domains:
soft partition of unity applies unchanged to a 1D signal, with each site
parameterized by a position, value, log-temperature, and radius
(4 parameters, versus the 9 used in 2D). We compare \name to two 1D
baselines at approximately matched parameter budget (64 primitives, 256 parameters):
(i)~front-to-back Gaussian splatting (position, $\sigma$, opacity, value per
splat), which blends overlapping Gaussian kernels; and (ii)~SIREN, a small
sinusoidal coordinate MLP that encodes the signal through global frequencies.

Figure~\ref{fig:1d_signal} reports results on three characteristic targets.
Gaussian splatting dominates the smooth sine ($79.6$~dB) because a Gaussian
basis is near-optimal for band-limited interpolation, but its fixed-shape
kernels round off the sharp step ($47.3$~dB): it cannot model a discontinuity
\emph{directly}. SIREN handles continuous variation through its global
sinusoidal basis, but approximates discontinuities only by truncating their
frequency content, producing Gibbs-like ringing around the step ($43.1$~dB).
\name, in contrast, exploits its learnable per-site temperature to sharpen the
soft partition \emph{exactly} at the discontinuity while staying smooth in the
flat regions, leading on both the step ($57.2$~dB) and multi-frequency
($42.0$~dB) targets. This mirrors the same mechanism that aligns \name cells
with object boundaries in 2D, confirming that temperature-controlled ownership
---not the site budget or dimensionality---is what lets \name represent sharp
structure without sacrificing smoothness.
}

\section{Conclusion}
We presented Soft Anisotropic Diagrams (\name), an explicit and differentiable image representation based on a soft anisotropic additively weighted Voronoi (Apollonius-style) partition of the image plane. Pixels are rendered as a temperature-controlled softmax blend over a small per-pixel top-$K$ subset of adaptive sites, which keeps optimization well-conditioned while making spatial ownership explicit and allowing boundaries to sharpen where the content demands it. In \name, we maintain the per-query top-$K$ map under the same shading score and update it with a jump-flood-inspired propagation scheme with stochastic injection, enabling GPU-friendly, fixed-size local computation. Combined with a GPU-first training pipeline (gradient-weighted initialization, Adam optimization, and adaptive densification/pruning), this substantially reduces per-instance encoding cost without sacrificing quality. Across standard benchmarks, \name consistently outperforms Image-GS and Instant-NGP at matched bitrate. We also perform ablation studies to demonstrate the effectiveness of our method.

\paragraph{Limitations and future work.}
\name relies on maintaining accurate per-pixel top-$K$ candidate sets via
propagation; while effective in practice, imperfect or stale lists can reduce quality at very low budgets, after aggressive densification/pruning, or during abrupt parameter changes. Moreover, the fastest rendering regime assumes cached top-$K$ lists; when the candidate map must be refreshed (e.g., after large edits or when re-rendering at a new resolution), multiple propagation passes and/or re-seeding are required and the update can dominate end-to-end rendering time, especially at high resolutions. While \name captures sharp boundaries well, very fine stochastic textures and highly irregular natural-image microstructure may still require larger site budgets (or temperature schedules) to avoid visible oversmoothing or residual grain. {We also note that the explicit structure of SAD is designed to enforce locality, rather than to guarantee exact edge alignment; in practice, boundary alignment with image structure is an emergent optimization outcome and can still degrade on thin structures, weak-contrast edges, or highly stochastic textures.} Finally, our implementation is optimized for GPU execution and bandwidth-efficient kernels, so absolute performance and the best configuration choices may vary across hardware and kernel configurations.

Promising directions include hierarchical and adaptive candidate maintenance---multi-resolution caches, resolution-aware reuse, and streaming-friendly refresh schedules---to reduce update cost while preserving quality at extreme budgets.
Another direction is richer per-site models beyond constant color (e.g., low-order local appearance, small learned decoders, or material cues where applicable). Equally important is improving the distance model itself: incorporating better-designed anisotropic metrics and/or learned distance functions could increase expressivity without proportionally increasing site count. We also expect better initialization to further reduce optimization time: instead of gradient-based heuristics, one could use a pre-trained prior to propose an initial site layout and attributes, in the spirit of deep priors used for fast primitive initialization (e.g., \cite{wang2025fast}).

Beyond 2D images, extending \name to 3D (e.g., anisotropic cells for volumetric or surface representations) and to 2D/3D inverse problems could broaden its utility in differentiable rendering and physics. Finally, the explicit sites and induced adjacency suggest ML uses such as tokenization/encoding of images into structured primitives and using \name as an explicit decoder or generative domain in place of implicit coordinate MLPs.

\paragraph{Ethical and societal impact.}
Our method targets compact, differentiable \emph{image} representation with fast per-image fitting, and currently does not add new capabilities for content generation or identity inference beyond standard neural primitive-based codecs. As with any image representation technique, it may be misused to store or transmit sensitive imagery or to support downstream manipulation at scale, raising privacy (lack of consent), IP, and deceptive-editing concerns. We therefore position it as a drop-in representation for user-provided images, encourage responsible data practices (consent, access control, licensing compliance), and emphasize transparent, reproducible implementation. Societally, improved rate--distortion and faster encoding can reduce bandwidth/storage costs and benefit resource-constrained deployment.

\bibliographystyle{ACM-Reference-Format}
\bibliography{references.bib}

\clearpage
\appendix
\renewcommand\thefigure{\Alph{section}\arabic{figure}}    
\renewcommand\thetable{\Alph{section}\arabic{table}}
\section{Appendix}

\section{Implementation Details}
\label{supp_sec:implementation}

\subsection{Parameters and Constants}
Table~\ref{tab:impl_params} lists the runtime parameters referenced in this
section, while Table~\ref{tab:impl_consts} lists fixed constants shared across
backends.

\begin{table}[H]
  \centering
  \small
  \begin{tabular}{l l}
    \hline
    Parameter & Meaning \\
    \hline
    $K$ & Top-$K$ candidate list size (we use $K=8$) \\
    $s_{\mathrm{grid}}$ & Candidate grid downscale factor \\
    $f_{\mathrm{cand}}$ & Candidate update period (iterations per update) \\
    $n_{\mathrm{cand}}$ & Passes per candidate update \\
    \hline
  \end{tabular}
  \caption{Runtime parameters used by the implementation.}
  \label{tab:impl_params}
\end{table}

\begin{table}[H]
  \centering
  \small
  \begin{tabular}{l l}
    \hline
    Constant & Meaning \\
    \hline
    Grad-quant scale ($10^6$) & Scale for integer gradient accumulation \\
    Tile hash size (256) & Number of slots in the per-tile hash table \\
    Max probes (8) & Linear-probing cap per hash insertion \\
    Empty key (0xffffffff) & Sentinel for unused hash slots / invalid IDs \\
    \hline
  \end{tabular}
  \caption{Fixed constants shared across backends.}
  \label{tab:impl_consts}
\end{table}

\subsection{Data Layout and Packing}
\paragraph{Site buffer.}
Training uses full-precision float32 site parameters. Each site stores
the 10 semantic values $(x, y, \log \tau, r, c_r, c_g, c_b, a_x, a_y, a)$.
The sentinel for inactive sites is $x<0$ (we set position to $(-1,-1)$), and
all kernels skip inactive entries. This layout is shared across Metal, CUDA,
and WebGPU.

\paragraph{Packed storage for evaluation.}
For evaluation we render from a 16-byte packed site format that encodes all
10 semantic values. The layout is four 32-bit words:
\textbf{(w0)} position $x,y$ as 15-bit UNORM each (scaled by image width/height)
plus an active flag in the high bits; \textbf{(w1)} color $r,g,b$ as 11/11/10-bit
UNORMs; \textbf{(w2)} $(\log\tau, r)$ as 16/16-bit UNORMs using per-image
min/scale; \textbf{(w3)} lower 16 bits = anisotropy direction as a 16-bit
UNORM angle code over $[-\pi,\pi]$, upper 16 bits = $a$ as float16. The per-image quantization
parameters (min/scale for $\log\tau$, $r$, and color channels) are stored once
alongside the packed array. In our evaluation this packed representation does
not change PSNR, while reducing memory bandwidth substantially. Candidate
updates use a separate 16-byte half2-packed format for speed.

\paragraph{Top-$K$ candidate list.}
We store the per-pixel top-$K$ indices in two 4-channel 32-bit unsigned integer
textures, each texel containing four site IDs. With $K=8$, the first texture
stores IDs $0$--$3$ and the second stores IDs $4$--$7$. The invalid sentinel is
0xffffffff. This design yields coalesced loads for both rendering and gradients.

\paragraph{Packed candidate sites.}
For candidate updates we use a compact per-site representation to reduce
bandwidth. Each site is packed into 16 bytes (four 32-bit words) using
half-precision pairs (two float16 per word): position, $(\log\tau, r)$,
anisotropy direction, and $(a, 0)$. The packed buffer is regenerated
after densification and used by the candidate update kernel.

\subsection{Candidate Field and Jump Schedule}
\paragraph{Downscaled candidate grid.}
We maintain a candidate grid of size
$\lceil W / s_{\mathrm{grid}} \rceil \times \lceil H / s_{\mathrm{grid}} \rceil$ with downscale $s_{\mathrm{grid}}$. The mapping
from image coordinates to candidate cells is integer (no interpolation), while
candidate evaluation uses the centered UV of the candidate cell to avoid
systematic bias.

\paragraph{Step encoding.}
The jump schedule is encoded in a single 32-bit step parameter:
the lower 16 bits store the step index, and the upper 16 bits store the jump
distance. This allows a compact parameter buffer and deterministic per-pass
randomization (xorshift state is seeded by the step index and pixel ID).

\paragraph{JFA prepass.}
When a full refresh is needed (e.g., for rendering), we run a seed pass and
$\lceil \log_2 \max(W,H) \rceil$ flood passes. Each flood pass samples a 3$\times$3
neighborhood at the current step size and writes the four closest sites into
the first candidate texture. This prepass is optional: the candidate field can
be built from VPT updates alone, while JFA mainly helps when reinitializing or
after large edits.

\paragraph{Candidate update frequency.}
We expose the candidate update frequency and number of passes per update.
In our experiments, updating the candidate field less frequently (e.g., once
every 8--16 iterations) did not measurably change the final PSNR, while further
reducing training time.

\subsection{Gradient Computation and Reduction}
\paragraph{Per-pixel gradients.}
Gradients are computed from the softmax weights of the top-$K$ list. We use
the standard max-subtracted softmax for numerical stability and compute
derivatives with respect to position, $\log \tau$, $r$, color, anisotropy
direction, and $a$ (anisotropy scale). Invalid sites are skipped and all
NaNs are guarded.

\paragraph{Quantized vs. float atomics.}
WebGPU uses 32-bit integer gradient buffers with a fixed scale (grad-quant
scale in Table~\ref{tab:impl_consts}) because floating-point atomics are not
universally available. Metal and CUDA use native float atomics for the same
gradients. In all backends, the Adam update converts the accumulated
values back to floating point and normalizes by the number of pixels.

\paragraph{Threadgroup hash reduction.}
We implement a fixed-size, per-tile hash table sized to the 16$\times$16
workgroup (256 slots). Each slot stores a site ID key and 11 accumulators
(10 gradients + removal-delta). Slots are initialized to the empty
sentinel 0xffffffff with per-thread strided clearing. Keys are hashed
with a multiplicative hash ($key\times2654435761$), and we use linear
probing with a hard cap of 8 probes. Insertion is via atomic compare-and-swap
on the key; if a probe hits an existing key, we accumulate into that slot.
If no slot is found within the probe bound, we fall back to global atomics for
that site.
After all pixels in the tile are processed, we synchronize and have each thread
flush a strided subset of slots: for every occupied slot, we perform one global
atomic per parameter (and removal-delta when enabled). In Metal and WGSL, the
threadgroup accumulators are stored as scaled 32-bit integers (grad-quant scale
in Table~\ref{tab:impl_consts}) and dequantized on flush; in CUDA they are stored
as float in shared memory. This yields at most one global atomic per site per
tile and keeps the hash-table overhead bounded and deterministic.

\subsection{Densification and Pruning}
\paragraph{Densification scores.}
A statistics pass accumulates per-site mass, energy, and weighted second
moments $(w, wx, wy, wxx, wxy, wyy)$ from the per-pixel residuals. We score
each site as $\mathrm{energy} / \mathrm{mass}^{\alpha}$ and use a radix sort
over key-value pairs to select the top candidates for splitting.

\paragraph{Split kernel.}
Each selected site is split into two children. When sufficient statistics are
available, we estimate the dominant axis from the weighted covariance; if not,
we fall back to a local Sobel-like gradient around the site. Children are
offset along this axis, inherit color from the target image at their positions,
and slightly reduce $\log \tau$ and $r$ (e.g., $r \leftarrow 0.85\,r$).
The Adam state for both parent and child is reset to zero.

\paragraph{Pruning.}
We compute a per-site removal score from the removal-delta accumulator,
sort the scores, and mark the lowest-ranked sites as inactive by setting
position to $(-1,-1)$. Inactive sites are ignored by all kernels.

\subsection{Tau Diffusion}
We optionally smooth $\log \tau$ gradients by averaging over the local candidate
neighborhood at each site. This is implemented as a per-site Jacobi update that
mixes the raw gradient with neighboring gradients using a scalar $\lambda$.

\subsection{Dispatch Configuration}
Table~\ref{tab:dispatch_sizes} summarizes the workgroup sizes. Image-space
kernels (render, candidate update, gradients, stats, JFA flood/clear) use
16$\times$16 groups. Per-site kernels use 1D groups of 64 threads (JFA seed)
or 256 threads (Adam, split, prune, pack, tau diffusion).

\begin{table}[t]
  \centering
  \small
  \begin{tabular}{l c}
    \hline
    Kernel family & Workgroup size \\
    \hline
    Candidate init/clear/update, render, gradients, stats & $16\times16$ \\
    JFA seed & $64\times1$ \\
    Pack / Adam / split / prune / tau diffusion & $256\times1$ \\
    \hline
  \end{tabular}
  \caption{Typical dispatch sizes used across backends.}
  \label{tab:dispatch_sizes}
\end{table}

\subsection{Backend Notes}
The Metal, CUDA, and WebGPU implementations share identical math, memory
layout, and candidate update logic. WebGPU uses a shared WGSL source (also used
by the JS viewer), while Metal and CUDA translate the same kernels into MSL and
CUDA C++. The main differences are binding models and buffer creation; all
buffers are kept in GPU memory and only site I/O crosses the host boundary.
We avoid backend-specific features in the training path to keep behavior
consistent across devices.

\end{document}